\documentclass[useAMS,referee,usenatbib]{biom}
\usepackage{bm}
\usepackage{graphicx}
\def\bSig\mathbf{\Sigma}

\usepackage{amsmath}
\usepackage{amsfonts}
\usepackage{amssymb}
\usepackage{color}
\usepackage{algorithm}
\usepackage{algpseudocode}
\usepackage{algorithmicx}
\usepackage{multirow}
\usepackage{mathtools}
\usepackage{hyperref}
\usepackage{url}
\newcommand{\mbf}{\mathbf}

\author{Hongju Park$^{1,2}$, 
Shuyang Bai$^{3}$, Zhenyao Ye$^{1,2}$, Hwiyoung Lee$^{1,2}$, Tianzhou Ma$^{4}$, Shuo Chen$^{1,2,*}$\email{shuochen@som.umaryland.edu}\\
$^{1}$Maryland Psychiatric Research Center, School of Medicine, University of Maryland \\
$^{2}$The University of Maryland Institute for Health Computing (UM-IHC)\\
$^{3}$Department of Statistics, University of Georgia\\
$^{4}$Department of Epidemiology and Biostatistics, School of Public Health, University of Maryland
}
\title[Graph Canonical Correlation Analysis]{Graph Canonical Correlation Analysis}

\begin{document}

\label{firstpage}
\begin{abstract}
Canonical correlation analysis (CCA) is a widely used technique for estimating associations between two sets of multi-dimensional variables. Recent advancements in CCA methods have expanded their application to decipher the interactions of multiomics datasets, imaging-omics datasets, and more. However, conventional CCA methods are limited in their ability to incorporate structured patterns in the cross-correlation matrix, potentially leading to suboptimal estimations. To address this limitation, we propose the graph Canonical Correlation Analysis (gCCA) approach, which calculates canonical correlations based on the graph structure of the cross-correlation matrix between the two sets of variables. We develop computationally efficient algorithms for gCCA, and provide theoretical results for finite sample analysis of best subset selection and canonical correlation estimation by introducing concentration inequalities and stopping time rule based on martingale theories. Extensive simulations demonstrate that gCCA outperforms competing CCA methods. Additionally, we apply gCCA to a multiomics dataset of DNA methylation and RNA-seq transcriptomics, identifying both positively and negatively regulated gene expression pathways by DNA methylation pathways. 
\end{abstract}
%  Please place your key words in alphabetical order, separated
%  by semicolons, with the first letter of the first word capitalized,
%  and a period at the end of the list.
%
%\keywords{Canonical Correlation Analysis, Best Subset Selection, Joint High-dimensional Data Analysis}


%  As usual, the \maketitle command creates the title and author/affiliations
%  display 

\maketitle

%  As usual, the \maketitle command creates the title and author/affiliations
%  display 

%  If you are using the referee option, a new page, numbered page 1, will
%  start after the summary and keywords.  The page numbers thus count the
%  number of pages of your manuscript in the preferred submission style.
%  Remember, ``Normally, regular papers exceeding 25 pages and Reader Reaction 
%  papers exceeding 12 pages in (the preferred style) will be returned to 
%  the authors without review. The page limit includes acknowledgements, 
%  references, and appendices, but not tables and figures. The page count does 
%  not include the title page and abstract. A maximum of six (6) tables or 
%  figures combined is often required.''

%  You may now place the substance of your manuscript here.  Please use
%  the \section, \subsection, etc commands as described in the user guide.
%  Please use \label and \ref commands to cross-reference sections, equations,
%  tables, figures, etc.
%
%  Please DO NOT attempt to reformat the style of equation numbering!
%  For that matter, please do not attempt to redefine anything!

\section{Introduction}
\label{sec:1}

Canonical Correlation Analysis (CCA) is one of the most widely used methods for exploring relationships between two sets of high-dimensional data \citep{hotellingrelations,yang2019survey,zhuang2020technical}. In biomedical data analysis, for example, CCA has been applied to study the coupling between structural and functional brain imaging variables in neuroscience research and to examine intercorrelated pathways between epigenetic and transcriptomic measures (i.e., multi-omics) in molecular biology research \citep{zhou2024fair,lee2024dcca}. In most CCA models, the objective is to identify linear combinations of the variables in each dataset, known as canonical variables, that are maximally correlated. However, traditional CCA has limited applications in high-dimensional data analysis. As the number of variables increases, canonical correlations tend to be inflated, leading to overly high estimates—similar to the inflation of R-squared values in linear regression when overfitting occurs. Furthermore, when the sample size is smaller than the number of variables, CCA cannot be computed due to the presence of non-invertible matrices \citep{le2009integromics}.

To address this limitation, various CCA methods in a sparse setup have been studied. \cite{witten2009extensions,witten2009penalized} developed CCA methods with penalties (i.e. $\ell_1$) imposed on canonical vectors. Following this work, \cite{tenenhaus2014variable} generalizes penalty functions of CCA for 3 or more data sets. Although these CCA methods with  $\ell_1$ penalization perform well to identify the maximum correlations between the two sets of multi-dimensional variables, they may not fully recover the sets of correlated variables in the two sets \citep{buhlmann2011statistics}. The partially recovered sets of correlated variables may be limited to revealing the systematic relationships between the two sets of variables. For example, in our motivating multi-omics dataset, we aim to investigate how DNA methylation variables regulate RNA expression. Existing CCA methods are limited in uncovering underlying intercorrelated pathways and estimating correlations between them.

To bridge the gap, we propose a graph Canonical Correlation Analysis (gCCA) model that simultaneously reveals the correlated variables and estimates the canonical correlation based on the graph patterns of the cross-correlation matrix of the two data sets. By leveraging the graph patterns, gCCA can identify the correlated modules/pathways in the two sets and estimate canonical correlations as a measure of association between the modules. This often provides more interpretable findings for biomedical research, for example, to extract systematically related multi-omics/imaging-omics data revealing correlated pathways. In addition, we theoretically demonstrate a minimum sample size condition that guarantees the performance of gCCA to identify all correlated variables between two datasets with high probability. Lastly, we establish a finite-sample guarantee that demonstrates a square-root convergence rate for canonical correlation estimation through the identification of associated variables.

%Additionally, we provide a theoretical demonstration of a minimum sample size condition that guarantees gCCA's ability to identify all correlated variables between two datasets with high probability.

%\tcb{I would like to emphasize that the estimation accuracy is constructed on the finite sample analysis. I think that ``a square-root convergence rate" means an asymptotic property, as my understanding. Could we have some other expression for the finite sample estimation accuracy? }

%\tcr{Ray: I see. How about just saying: we establish a finite-sample guarantee bound that implies a square-root convergence rate.}

%\tcr{(Ray: $p,q,p_0,q_0$ undefined yet. Notation $n_0\ge O(\cdots)$ may need to be revised: $Q=O(\cdots)$ usually denotes order of $Q$ does NOT exceed $\cdots$, and in general should NOT be written in a lower bound. How about $n\geq   C\max(p,q)^2/\min(p_0,q_0)^2$ for some constant $C>0$?)} 
%\tblue{
%$C\max(p,q)^2/\min(p_0,q_0)^2$ might not clearly represent the minimum sample size based on the one in Theorem 1, because it has some terms like log max(p,q).}  
%\tcr{Ray: I see, but still $n_0 = O(\max(p,q)^2/\min(p_0,q_0)^2)$ allows the possibility of $n_0=0$. One might have a vague description here, say $n$ should not be of lower order than $\max(p,q)^2/\min(p_0,q_0)^2$ up to logarithmic factor.}

%Even though this method with penalization shows good prediction performance, it does not guarantee the set of exact associated variables \citep{buhlmann2011statistics}. To obtain the set of exact associated variables, we need different constraints for variable selection.

\begin{figure}[h]
\centering
\includegraphics[width=0.8\textwidth]{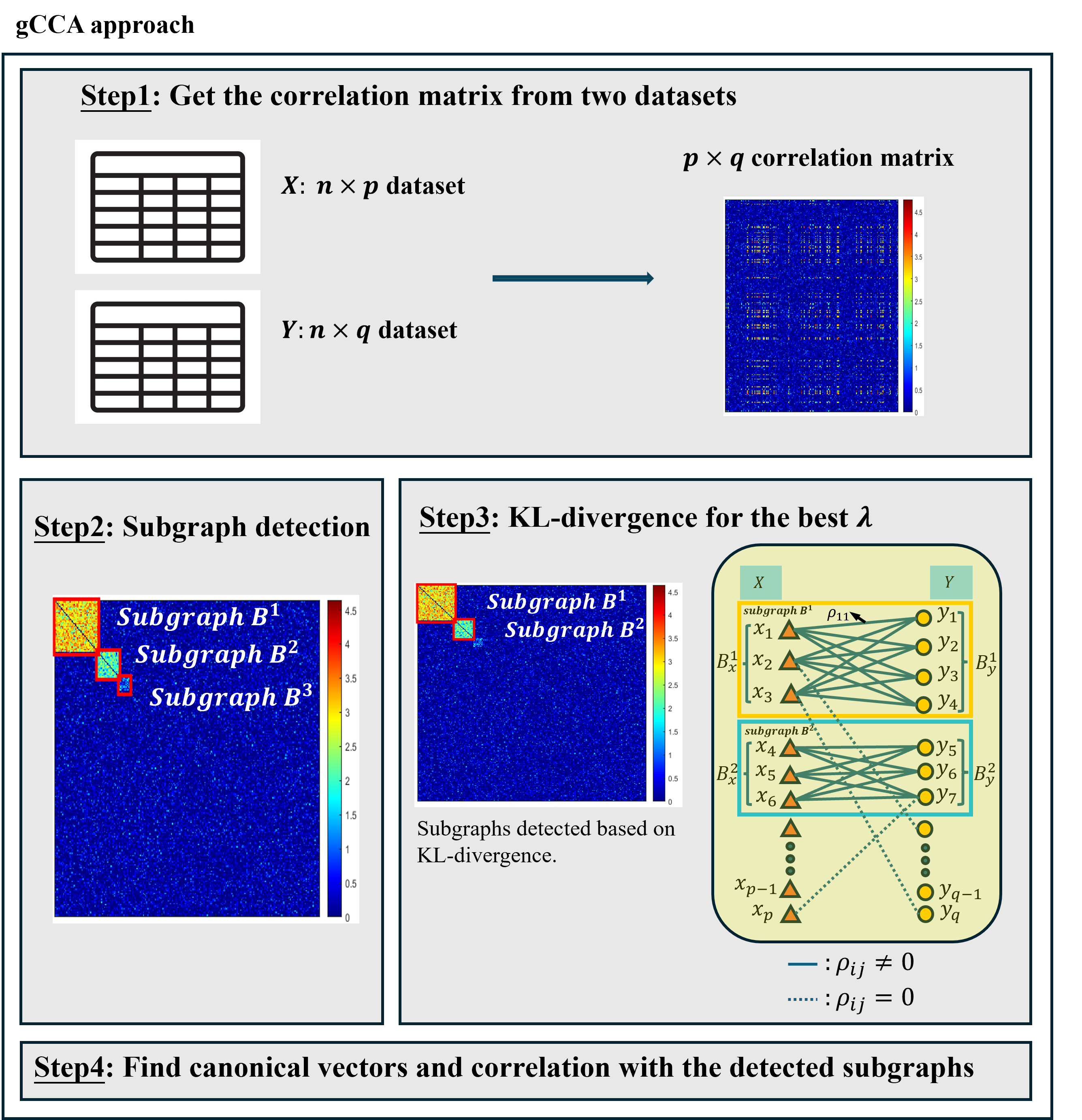}
\caption{Pipeline to detect associated variables for associations in two different data sets by gCCA. Step 1 shows the correlation matrix calculated from two joint datasets. Step 2 illustrates the subgraphs detected by the greedy algorithm for a specific tuning parameter $\lambda$. Step 3 showcases the two optimal subgraphs based on the optimal tuning parameter and the connections of variables in and outside the subgraphs. Lastly, Step 4 is the calculation of canonical vectors and correlation.}
\label{fig:1}
\end{figure}

Figure \ref{fig:1} provides an overview of the gCCA method. The procedure begins with calculating the sample cross-correlation matrix between two sets of variables from the same subjects. Following this, gCCA is applied with a set of tuning parameter values to identify graph patterns, which we refer to as subgraphs with strong associations throughout this paper. The optimal tuning parameter is then selected objectively. The extracted subgraphs allow us to interpret the interacted modules or pathways within the two sets of variables. Finally, we compute the canonical vectors and correlations to quantify the strength of associations between the two data sets based on the extracted graph patterns.

The organization is as follows. In Section \ref{sec:2}, we introduce the background of CCA, graph Canonical Correlation Analysis (gCCA) model, and a greedy algorithm for gCCA to detect subgraphs in which variables have strong associations. In addition, we introduce a set of assumptions and theoretical results based on the assumption in Section \ref{sec:3}. Next, we evaluate gCCA using synthetic datasets in Section \ref{sec:4} and apply it to the motivating dataset in Section~\ref{sec:5} to study the interactions between DNA methylations and transcriptomics in patients with cancer. The paper concludes with a discussion.

\section{Methods}
\label{sec:2}

\subsection{Data structure}

In this work, we analyze the associations of two datasets ${\bf X} \in \mathbb{R}^{n\times p}$ and ${\bf Y} \in  \mathbb{R}^{n\times q}$, where the $s$-th rows of $\mbf{X}$ and $\mbf{Y}$ represent the measurements of $s$-th subject for $s=1,\dots,n$, denoted by $X^{(s)} = (X_1^{(s)},\dots,X_p^{(s)})$ and $Y^{(s)} = (Y_1^{(s)},\dots,Y_q^{(s)})$, which are samples of random vectors $(X,Y)$.  Suppose without loss of generality that the datasets each have been centered and scaled to have sample mean zero and sample second moment 1 for each column component of ${\bf X}$ and ${\bf Y}$. In this work, we do not impose specific distributional assumptions on $X$ and $Y$, but instead conduct our analysis under a set of general assumptions, which will be formally introduced and discussed in the next sections.

\subsection{Background}

CCA is a statistical method used to explore the relationships between two datasets. CCA considers the following maximization problem: $$\text{max}_{a,b} (a^\top {\bf X}^\top {\bf Y} b)~\text{subject~to}~a^\top {\bf X}^\top {\bf X} a \leq 1~\text{and}~b^\top {\bf Y}^\top {\bf Y} b \leq 1,$$
where the vectors $a$ and $b$ and the correlation are said to be canonical vectors and canonical correlation if they attain the above maximization.
% \tcr{(Ray: what is the precise meaning of centered and scaled? Each row of $\mathbf{X}$ has sample mean zero and sample second moment 1?) HP: I clarified it.} \tcr{(Ray: may be helpful to give a citation for the formula below?)}
%\begin{eqnarray*}
%  \text{max}_{a,b} (a^\top {\bf X}^\top {\bf Y} b)~\text{subject~to}~a^\top {\bf X}^\top {\bf X} a \leq 1~\text{and}~b^\top {\bf Y}^\top {\bf Y} b \leq 1,
%\end{eqnarray*}
%which is an approximate sample version of \eqref{eq:cano} \citep{witten2009extensions,witten2009penalized,zhou2024fair,lee2024dcca}. A maximizer for the above problem is 
%written in the closed form 
%where the solutions are given by $\hat{a} = u_1({\bf X}^\top {\bf Y})$, $\hat{b} = v_1({\bf X}^\top {\bf Y})$, $u_1(M)$ and $v_1(M)$ are the left and right singular vectors corresponding to the first singular value of a matrix $M$, respectively. 
%The estimated canonical correlation is then calculated as follows: $\widehat{\rho} = \hat{a}^\top {\bf X}^\top {\bf Y} \hat{b}/ \sqrt{(\hat{a}^\top {\bf X}^\top {\bf X} \hat{a})( \hat{b}^\top {\bf Y}^\top {\bf Y} \hat{b})}$.

%\textit{Canonical Correlation Analysis with Regularization }

In the classical canonical correlation analysis, the canonical vectors $a$ and $b$ include nonzero loadings for all $X$ and $Y$ variables. However, in a high-dimensional setting with $p,q\gg n$, the goal is to identify which subsets of $X$ are associated with subsets $Y$ and estimate the measure of associations, as the canonical correlation with the full dataset is overly high due to estimation bias caused by overfitting. To ensure the sparsity, shrinkage methods are commonly used. For example, \cite{witten2009penalized} propose sparse canonical correlation analysis (sCCA). The criterion of sCCA can be in general expressed as follows:
%In the classical canonical correlation analysis, all the elements of $X$ and $Y$ are assumed to be associated with the canonical vectors $a$ and $b$. However, in a high-dimensional setting with $p,q\gg n$, to obtain meaningful results, it is preferable to assume that there are associated relationships between only a portion of the elements of $X$ and $Y$. Accordingly, we consider a sparse setup where only a subset of the elements of $X$ and $Y$ is related to the canonical vectors. This problem can be represented as the setup where only a small portion of the components of $a$ and $b$ are non-zero, which is written as \eqref{eq:l0} in the next subsection. However, the estimation of non-zero subsets is computationally intractable due to the cardinality constraint, which renders it NP-hard \citep{natarajan1995sparse,bertsimas2016best}. Accordingly, \cite{witten2009penalized} propose sparse canonical correlation analysis (sCCA). The criterion of sCCA can be in general expressed as follows:
\begin{eqnarray*}
\max_{a,b} a^\top {\bf X}^\top {\bf Y} b~\text{subject to}~a^\top {\bf X}^\top {\bf X} a \leq 1,~b^\top {\bf Y}^\top {\bf Y} b \leq 1,~P_1(a) \leq k_1,~P_2(b) \leq k_2,
\end{eqnarray*}
where $P_1$ and $P_2$ are convex penalty functions for penalization for $a$ and $b$ with positive constants $k_1$ and $k_2$, respectively. A representative penalty function is a $\ell_1$ penalty function such that $P_1(a) = \|a\|_1$ and $P_2(b) = \|b\|_1$. sCCA imposes zero loadings in canonical vectors and thus only selects subsets of correlated $X$ and $Y$. However, sCCA methods may neither fully recover correlated  $X$ and $Y$ pairs nor capture the multivariate-to-multivariate linkage patterns (see Figure \ref{fig:3}) because the $\ell_1$ shrinkage tends to select only a small subset from the associated variables of $X$ and $Y$.

%\tcr{(Ray: it is not clear to me what $X_0$ and $Y_0$ mean.)}

\subsection{Graph Canonical Correlation Analysis}

%Data notation We propose the gCCA model for two sets of variables X, Y. subject variable, $x_{i}^s$, $y_{j}^s$ $i=1, \cdots p$,$j=1, \cdots q$, $s=1, \cdots n$. 
To capture the systematic correlations between $X$ and $Y$ and estimate their strength, we present the gCCA method to solve the following objective function:
\begin{eqnarray}
  \max_{a,b} a^\top {\bf X}^\top {\bf Y} b~\text{subject~to}~a^\top {\bf X}^\top {\bf X} a \leq 1,~b^\top {\bf Y}^\top {\bf Y} b\leq 1 ~\text{and}~\|a\|_0\leq k_1,~\|b\|_0\leq k_2.
  \label{eq:l0}
\end{eqnarray}
where $\|\cdot\|_0$  is the $\ell_0$ norm and $k_1$ and $k_2$ are positive constants. The goal is to capture maximum correlations between $X$ and $Y$ with the minimal subsets of $X$ and $Y$ included. Therefore, gCCA can recover correlated components of $X$ and $Y$ and better estimate the canonical correlation while suppressing the false positive correlations. However, in practice, optimizing the gCCA objective function \eqref{eq:l0} is challenging due to the non-convexity and cardinality constraint, which renders it NP-hard \citep{natarajan1995sparse,bertsimas2016best}. Thus, we implement \eqref{eq:l0}  using a graph-based approach.

%However, the constraints in \eqref{eq:l0} as well as those in sCCA do not inherently capture the underlying relationships between $X$ and $Y$, making it difficult to interpret the associations based solely on the solution. To address this limitation, gCCA conceptualizes the correlation matrix as a bipartite graph $G=(U,V,E)$

We first define $G = (U, V, E)$ as a bipartite graph, where $U$ and $V$ are disjoint sets of nodes corresponding to the variables of $X$ and $Y$, respectively, and $E$ is the set of binary edges representing the presence of correlations between the variables of $X$ and $Y$. We utilize the concept of a bipartite graph to emphasize edges connecting $X$ and $Y$, rather than those within each disjoint node set.

The adjacency matrix ${\bf A} \in \mathbb{R}^{p\times q}$ represents the edges in the node sets $U$ and $V$. The adjacency matrix \( A = [A_{ij}] \) is a matrix-based representation of the edge set $E$. Each entry \( A_{ij} \) in the matrix is binary for the bipartite binary graph $G$ and has a relationship with the edge $e_{ij}$ between the node $u_i$ and $v_j$ as follows: $A_{ij} = 1$, if $e_{ij} \in E$ (edge exists), $A_{ij} = 0$, otherwise. Specifically, we let $A_{ij}=1$ if $\text{Corr}(X_i,Y_j) = \rho_{ij} \neq 0$ and  $A_{ij}=0$ otherwise. We have $|U|=p, |V|=q$ and $|E| = \sum_{i\in[p],i\in[q]}A_{ij}\leq pq$. 
%The degree of node $u_i$ is $\text{deg}_i=\sum_{j=1}^q e_{ij}$. When $X$ and $Y$ are systematically related to each other, nodes in $U$ and $V$ with high degrees are more likely to be connected (i.e., $\text{Prob} (e_{ij}=1)>\text{deg}_i \cdot \text{deg}_j/(p\cdot q)$ \tcr{(Ray: what randomness does ``$\text{Prob}$'' refer to? Randomness from uniformly sampling edges?)}). 
%We use a dense bipartite subgraph to characterize this pattern, $B_c=(U_c, V_c, E_c)$ for $c=1, \cdots, C$, where $P(e_{ij}=1|e_{ij} \in B_c)>P(e_{ij}=1|e_{ij} \notin B_c)$.
We use biclique (complete bipartite) subgraphs $B_c=(U_c, V_c, E_c)$ for $c=1, \cdots, C$ to characterize this pattern, where 
\begin{eqnarray*}
E_c = \{e_{ij}:A_{ij}=1~\text{for~all}~(i,j)\in U_c\otimes V_c\}, \end{eqnarray*}
is the edge set of the biclique subgraph with the disjoint node sets $U_c$ and $V_c$ and $I\otimes J$ for two sets $I$ and $J$ represents $\{(i,j):i\in I~\text{and}~j\in J\}$. Let $I_X=\cup_{c=1}^CU_c$ and $I_Y=\cup_{c=1}^CV_c$ be the index sets corresponding to the associated variables of $X$ and $Y$, respectively, which are denoted by $X_0$ and $Y_0$. 
Given $B=\{B_c\}_{c=1}^C$ and ${\bf X}_0$ and ${\bf Y}_0$, which are the subsets of ${\bf X}$ and ${\bf Y}$ corresponding to $X_0$ and $Y_0$, respectively, the objective function \eqref{eq:l0} can be re-written as
\begin{eqnarray}
    \max_{a,b}  a^\top {\bf X}_0^\top {\bf Y}_0 b~~~\text{subject to}~a^\top {\bf X}^\top_0 {\bf X}_0 a \leq 1,~b^\top {\bf Y}^\top_0 {\bf Y}_0 b\leq 1,\label{eq:l02}
\end{eqnarray}
with constraints $|I_X| \leq k_1$ and $|I_Y| \leq k_2$.
\subsection{Estimation}

In practice, neither biclique subgraphs $\{B_c\}_{c=1}^C$ nor corresponding variables $I_X$ and $I_Y$ are known. We estimate $U_c$ and $V_c$ (i.e., $I_X$ and $I_Y$) by the following objective function
%In practice, as the true sets of associated variables $I_X$ and $I_Y$ are unknown, we need to estimate them. To proceed, we propose an objective function %\tcr{(Ray: is $C$ known?)} \tblue{noted above that it is known}
\begin{eqnarray}
    f_\lambda(B,\varepsilon) = \sum_{c=1}^{C} \frac{ \sum_{(i,j): e_{ij} \in E_c} |R_{ij}^\varepsilon|}{(|U_c||V_c|)^\lambda},\label{eq:obj}
\end{eqnarray}
where $R_{ij}$ is a sample correlation of $X_i$ and $Y_j$ for a dataset with sample size $n$; $R_{ij}^\varepsilon = R_{ij}I(|R_{ij}|>\varepsilon)$; $\varepsilon\in (0,1)$ is a threshold value for cutting off absolute correlation values; $\lambda \in [0.5,1]$ is a tuning parameter. A greater $\lambda$ imposes a stricter penalty resulting in denser extracted subgraphs $\{B_c\}$ (i.e., higher density) yet covering a smaller number of edges with $|R_{ij}|>\varepsilon$. In \eqref{eq:obj}, $|U_c|$ and $|V_c|$ are equivalent to the $\ell_0$ norm for associated variables between $X$ and $Y$ in $\{B_c\}_{c=1}^C$, where $\sum_{c=1}^C |U_c|=\|a\|_0$ and $\sum_{c=1}^C |V_c|=\|b\|_0$.

%where $R_{ij}$ is a sample correlation of $X_i$ and $Y_j$; $R_{ij}^\varepsilon = R_{ij}I(|R_{ij}|>\varepsilon)$; $\lambda \in [0.5,1]$ is a tuning parameter; $\varepsilon\in (0,1)$ is a threshold value for cutting off absolute correlation values.  The objective function may be interpreted as the sum of $2\lambda$-dimensional density.
%of the matrix $|\mathbf{R}^{\varepsilon}|[I_X,I_Y]$, where $M[I_1,I_2]$ is the submatrix of $M$ with only $I_1=(i_1,i_2,\dots,i_{p_0})$-th rows and $I_2=(i_1',i_2',\dots,i_{q_0}')$-th columns for a matrix $M$
%For example, by letting $|U_c|=|V_c|=d$, $C=1$, and $\varepsilon=0$ for ease of presentation,  the value of the objective function with $\lambda=0.5$ ($\sum_{i,j}|R_{ij}^\varepsilon|/d$) indicates the average sum of all edges' weights (absolute correlation) connected to a randomly chosen node from a subgraph. In addition, for $\lambda=1$, the value of the objective function ($\sum_{i,j}|R_{ij}^\varepsilon|/d^2$) is the average edge weight (absolute correlation) of randomly chosen two components, where one is from $U_c$ and the other is from $V_c$. %\tcr{(Ray: I am having trouble understanding ``row or column in a subgraph''. There are only nodes and edges in a graph.)}\tcr{(Ray: I am struggling with the explanation of ``2-dimensional density'' here.)}

%we can consider the following optimization problem as a surrogate of \eqref{eq:l0}:

By implementing \eqref{eq:obj}, we estimate a canonical correlation using estimated $\widehat{I}_X$ and $\widehat{I}_Y$ from $\{\widehat{B}_c\}_{c=1}^C$ as follows:
\begin{eqnarray}
\widehat{a} = u_1({\bf X}[,\widehat{I}_X]^\top {\bf Y}[,\widehat{I}_Y])~~\text{and}~~\widehat{b} = v_1({\bf X}[,\widehat{I}_X]^\top {\bf Y}[,\widehat{I}_Y]),\label{eq:cvec}
\end{eqnarray}
where $u_1(M)$ and $v_1(M)$ are the left and right singular vectors corresponding to the first singular value of a matrix $M$, respectively; for a matrix $M \in \mathbb{R}^{n\times q}$ and integer set $I =(i_1,i_2,\dots,i_{q_0}) \subset [q]$, $M[,I]$ represent the submatrix of $M$ with $(i_1,i_2,\dots,i_{q_0})$-th columns. The estimated canonical correlation is
\begin{eqnarray}
    \widehat{\rho}_c = \frac{\widehat{a}^\top {\bf X}[,\widehat{I}_X]^\top {\bf Y}[,\widehat{I}_Y] \widehat{b}}{\sqrt{(\widehat{a}^\top {\bf X}[,\widehat{I}_X]^\top {\bf X}[,\widehat{I}_X] \widehat{a})(\widehat{b}^\top {\bf Y}[,\widehat{I}_Y]^\top {\bf Y}[,\widehat{I}_Y] \widehat{b}})}.\label{eq:ccor}
\end{eqnarray}
The details of the estimation of $I_X$ and $I_Y$ will be described as follows.

\textit{Greedy algorithm for \eqref{eq:obj}: }
We first implement the main gCCA objective function \eqref{eq:obj} in two steps: $\{\widehat{B}_c \}_{c=1}^C$ estimation by (3)  and canonical correlation estimation by \eqref{eq:cvec} and \eqref{eq:ccor}. Since the computations of \eqref{eq:cvec} and \eqref{eq:ccor} are straightforward, we focus on the implementation of the optimization of $f_\lambda(B,\varepsilon)$ in \eqref{eq:obj} with respect to $B$ using a greedy algorithm.

Algorithm \ref{algo1} outlines a pseudo-code for the greedy algorithm. This algorithm begins with two data sets ${\bf X}$ and ${\bf Y}$, which is centered and standardized, threshold $\varepsilon$, and values of tuning parameter $\bm{\lambda} = (\lambda_1,\lambda_2,\dots,\lambda_{g_0})$. Let $\circ$ be the Hadamard product (i.e., element-wise product) between matrices of the same dimensions,  and  for a matrix ${\bf M}$,  we define 
%\textit{Greedy algorithm for gCCA.} Here, we introduce a greedy algorithm that iteratively selects associated components for subgraphs based on the objective function defined in \eqref{eq:obj}. Table \ref{algo1} presents the pseudo-code for %\tcr{(Ray: why capitalize ``G''? Unless the name of the algorithm is called ``Greedy'',  which would have been a bit too unspecific.)} 
%the greedy algorithm applied to gCCA. The greedy algorithm starts with two data sets ${\bf X}$ and ${\bf Y}$, which is centered and standardized, threshold $\varepsilon$, and values of tuning parameter $\bm{\lambda} = (\lambda_1,\lambda_2,\dots,\lambda_{g_0})$. Let $\circ$ be the Hadamard product (i.e., element-wise product) between matrices of the same dimensions,  and  for a matrix $M$,  
\begin{eqnarray*}
(I({\bf M}>\varepsilon))_{ij} = \left\{\begin{array}{@{}lr@{}}
    1, & \text{if}~M_{ij} > \varepsilon,\\
    0, & \text{otherwise}.                       
    \end{array}\right.    
\end{eqnarray*}

We define the truncated absolute sample correlation matrix, denoted by $|\mathbf{R}^{\varepsilon}| \in \mathbb{R}^{p \times q}$, as follows: 
\begin{eqnarray*}
|\mathbf{R}^{\varepsilon}| = |\mathbf{R}| \circ I(|\mathbf{R}| > \varepsilon),    
\end{eqnarray*}
where  $\mathbf{R} = \mathbf{X}^\top \mathbf{Y}$  is the sample correlation matrix with a sample of size $n$. We initialize the process with $\mathbf{R}^{1,1} = |\mathbf{R}^{\varepsilon}|$, $J^{1,1}_X=[p]$ and $J^{1,1}_Y=[q]$. Let $\mathbf{R}^{c,t}$ denote the \textit{active matrix} and $J^{c,t}_X$ and $J^{c,t}_Y$ do the \textit{active sets} of rows $(X)$ and columns $(Y)$ of the matrix $|\mathbf{R}^\varepsilon|$, respectively, at time $t$ for the $c$-th subgraph. The matrix $\mathbf{R}^{c,t}$ is defined as $\mathbf{R}^{c,t} = |\mathbf{R}^{\varepsilon}|[J^{c,t}_X,J^{c,t}_Y]$. %where $J^{c,t}_X \subseteq [p]$ and $J^{c,t}_Y \subseteq [q]$ are the \textit{active sets} representing the sets of active rows and columns, respectively, at time $t$ for the $c$-th subgraph.

\begin{figure}[h]
\centering
\includegraphics[width=1\textwidth]{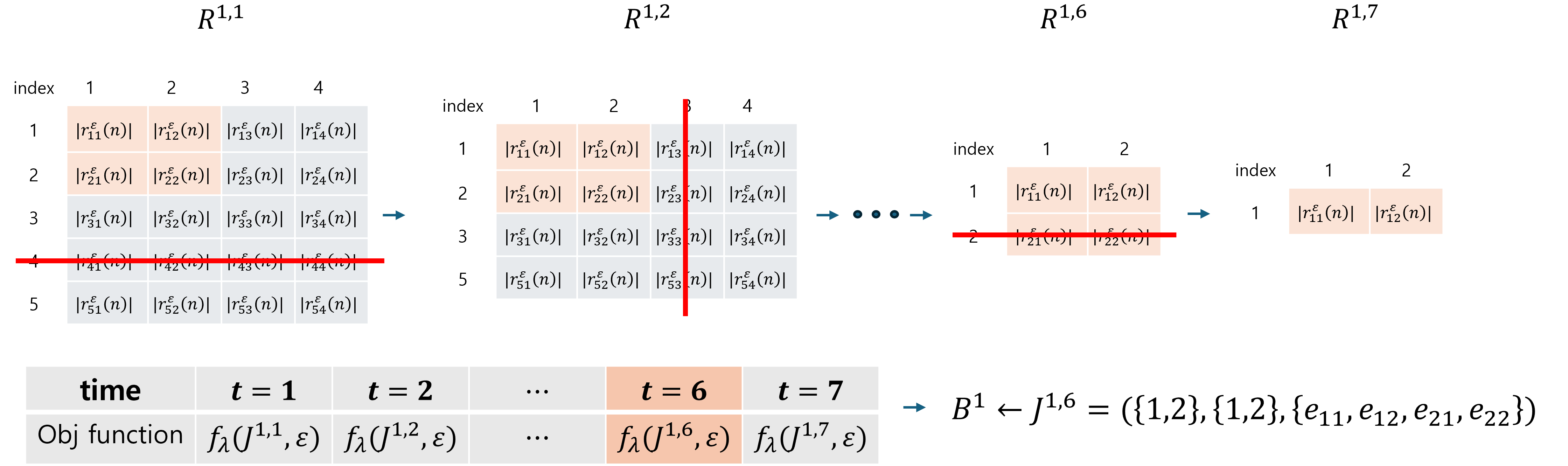}
\caption{Row and column exclusion process by the greedy algorithm under the presence of a subgraph (size: 2 by 2) in a graph with 5 rows and 4 columns. Red and gray cells represent (unknown) associated and irrelevant variables, respectively. Solid red lines indicate the exclusion of the row or column with the lowest row or column mean among all active rows and columns. The table shows that the objective function is maximized at $t=6$ and thereby $J^{1,6} = (\{1,2\}, \{1,2\}, \{e_{11},e_{12},e_{21},e_{22}\} )$ with $A_{ij}=1$ for all $(i,j)$ is considered the extracted biclique subgraph. }
\label{fig:greedy}
\end{figure}

Define $u^{c,t} = \text{vec}(J_X^{c,t})$ and $v^{c,t} = \text{vec}(J_Y^{c,t})$, where $\text{vec}(I)$ represent the vectorization of $I$ in the increasing order for a set $I$. For a given $\lambda_g$, at time $t$ for the $c$-th subgraph extraction, we calculate the row and column means of the active matrix, denoted by $\text{r.means}(\mathbf{R}^{c,t})$ and $\text{c.means}(\mathbf{R}^{c,t})$, such that 
$$
\text{r.means}(\mathbf{R}^{c,t})=   \left(\frac{1}{| J_Y^{c,t}|}\sum_{j\in J_Y^{c,t}}|R_{u^{c,t}_1j}^\varepsilon|,\frac{1}{| J_Y^{c,t}|}\sum_{j\in J_Y^{c,t}} |R_{u^{c,t}_2j}^\varepsilon|,\dots,\frac{1}{| J_Y^{c,t}|}\sum_{j\in J_Y^{c,t}}|R_{u^{c,t}_{|J_X^{c,t}|}j}^\varepsilon|\right)$$ and  $$\text{c.means}(\mathbf{R}^{c,t})= \left(\frac{1}{| J_X^{c,t}|}\sum_{i\in J_X^{c,t}}|R_{iv^{c,t}_1}^\varepsilon|,\frac{1}{| J_X^{c,t}|}\sum_{i\in J_X^{c,t}} |R_{iv^{c,t}_2}^\varepsilon|,\dots,\frac{1}{| J_X^{c,t}|}\sum_{i\in J_X^{c,t}}|R_{iv^{c,t}_{|J_Y^{c,t}|}}^\varepsilon|\right),
$$ 
where $R_{ij}^\varepsilon$'s are as in \eqref{eq:obj} and $|S|$ for a set $S$ represents the number of elements of $S$. A small row or column mean suggests that the corresponding sample correlation coefficients are closer to zero compared to others. %To standardize the sums of rows and columns, we compare the minimum row and column sum considering the sizes of rows and columns of $\mathbf{R}^{1,1}$. %the size ratio of which is denoted by \tcr{(Ray: there seems no need to introduce   $\text{s.ratio}$; one may simply compare  $\text{r.means}(\mathbf{R}^{1,1})_{\tau_1}/ \text{r.size}(\mathbf{R}^{1,1}) $ v.s.\ $ \text{c.means}(\mathbf{R}^{1,1})_{\phi_1})/\text{c.size}(\mathbf{R}^{1,1})$)}
%\begin{eqnarray*}
%    \text{s.ratio}(\mathbf{R}^{1,1}) = \frac{\text{c.size}(\mathbf{R}^{1,1})}{\text{r.size}(\mathbf{R}^{1,1})},
%\end{eqnarray*}
We let the indices of the row and column with the minimum sums based on the active matrix $\mathbf{R}^{c,t}$ denoted by $\tau_t$ and $\phi_t$, respectively. We exclude row $u^{c,t}_{\tau_t}$ or column $v^{c,t}_{\phi_t}$ (the indices based on the initial active matrix $|\mathbf{R}^\varepsilon|$) from our active sets and update our active set as follows:
if $\text{r.means}(\mathbf{R}^{c,t})_{\tau_t}  > \text{c.means}(\mathbf{R}^{c,t})_{\phi_t}$,
$$J_X^{c,t+1} \leftarrow J_X^{c,t},~J_Y^{c,t+1} \leftarrow J_Y^{c,t}\backslash \{v_{\phi_t}^{c,t}\},~\mathbf{R}^{c,t+1} \leftarrow |\mathbf{R}^{\varepsilon}|[J_X^{c,t+1},J_Y^{c,t+1}]$$
and otherwise
$$J_X^{c,t+1} \leftarrow J_X^{c,t}\backslash \{u_{\tau_t}^{c,t}\},~J_Y^{c,t+1} \leftarrow J_Y^{c,t},~\mathbf{R}^{c,t+1} \leftarrow |\mathbf{R}^{\varepsilon}|[J_X^{c,t+1},J_Y^{c,t+1}].$$
%where $\text{r.size}(\mathbf{R}^{1,1})$ and $\text{c.size}(\mathbf{R}^{1,1})$ represent the numbers of  rows and columns of $\mathbf{R}^{1,1}$. 
We repeat this exclusion process until only one row or column remain in our active sets. At the end of the exclusion process, we designate the biclique $J^{c,t}=(J_X^{c,t},J_Y^{c,t},E^{c,t})$ maximizing the contribution of the $c$-th subgraph to the summation in \eqref{eq:obj} as the $c$-th subgraph $B_c=(U_c,V_c,E_c)$. Figure \ref{fig:greedy} illustrates the row and column exclusion process by the greedy algorithm under the presence of a biclique subgraph.
% \tcr{(Ray: might be nice to include a graphical illusration, in the spirit of what you showed me using Excel, of this exclusion process.)}

For the ($c+1$)-th subgraph extraction, we repeat the exclusion process described above with new active sets $J_X^{c+1,1} = J_X^{c,1}\backslash U_c$, $J_Y^{c+1,1} = J_Y^{c,1}\backslash V_c$ and active matrix $\mathbf{R}^{c+1,1} = \mathbf{R}[J_X^{c+1,1},J_Y^{c+1,1}]$. This process is repeated $C$ times. After that, we record the extracted subgraphs $B^{\lambda_g} = \{ B_c\}_{c=1}^{C}$, which correspond to $\lambda_g$. We repeat this entire process for a set of tuning parameters $\{\lambda_1,\dots,\lambda_{g_0}\}$. Next, we choose the optimal tuning parameter value $\lambda_{g^\star}$ based on the Kullback-Leibler (KL) divergence, as described below.

The tuning parameter $\lambda$ has a significant effect on the extraction of subsets. A larger value of $\lambda$ typically results in denser and smaller subsets compared to those obtained with a smaller $\lambda$. To select the optimal $\lambda$, we use the KL divergence, which measures the distance between two distributions. To describe the procedure, consider two distributions $P_\lambda$ and $Q$ of $D_{ij} = I(|R_{ij}|>\varepsilon)$. %\tcr{(Ray: distribution of what? Probably better to introduce these after the description of $D_{ij}$)}. 
$P_\lambda$ is a distribution with subgraphs extracted based on the tuning parameter $\lambda$. $P_\lambda$ divides the correlation matrix into two distinct blocks: (i) the subgraphs $B^\lambda$ extracted by the tuning parameter $\lambda$, and (ii) the area outside the subgraphs. In block (i), $D_{ij}$ is more likely to be 1, whereas it is more likely to be 0 in block (ii). Specifically, $\{D_{ij}\}_{i,j}$ are assumed to have the following distribution:
\begin{eqnarray*}
    D_{ij} \sim \text{Bernoulli}(\pi_1)&~&\text{if}~(i,j)\in B_c~\text{for~some}~c \in [C]\\
    D_{ij} \sim \text{Bernoulli}(\pi_0)&~&\text{otherwise}.
\end{eqnarray*}
%\tcr{(Ray: how to justify i.i.d.\ assumption on $D_{ij}$? For example, $R_{12}$ and $R_{13}$ should be in general dependent since their calculations both involve dimension $1$ component. Check out ``Wishart distribution'' distribution of Gaussian covariance matrix.)}
In contrast, we consider a reference Bernoulli distribution $Q$ with no graph patterns between $X$ and $Y$ as follows:
\begin{eqnarray*}
    D_{ij} \sim \text{Bernoulli}(\pi)~\text{for all}~(i,j).
\end{eqnarray*}
The KL divergence between these two distributions  can be written as
\begin{eqnarray*}
D_{\mathrm{KL}}\left(P_\lambda \| Q\right) &=&
 \sum_{(i, j) \in \cup_{c=1}^{C} (U_c\bigotimes V_c)}\left(D_{i j} \pi_1 \log \frac{\pi_1}{\pi}+\left(1-D_{i j}\right)\left(1-\pi_1\right) \log \frac{\left(1-\pi_1\right)}{(1-\pi)}\right) \\
&+&  \sum_{(i, j) \notin \cup_{c=1}^{C} (U_c\bigotimes V_c)}\left(D_{i j} \pi_0 \log \frac{\pi_0}{\pi}+\left(1-D_{i j}\right)\left(1-\pi_0\right) \log \frac{\left(1-\pi_0\right)}{(1-\pi)}\right),
\end{eqnarray*}
where $\pi_0$ and $\pi_1$ are the sample mean of $D_{ij}$ outside and in the subgraphs, respectively; $\pi$ is the overall sample mean of $D_{ij}$. Then, the optimal tuning parameter is chosen so as to maximize the  KL divergence as follows:
\begin{eqnarray*}
    \lambda^\star=  \text{argmax}_\lambda D_{\mathrm{KL}}\left(P_\lambda \| Q\right).
\end{eqnarray*}

Here, we use the distribution $P_\lambda$ as an approximate distribution of $I(|R_{ij}|>\varepsilon)$ in subgraphs disregarding the dependencies between different $R_{ij}$. This approach, known as the variational method, is frequently employed for inference involving dependent Bernoulli variables, as the dependencies among Bernoulli random variables can lead to intractability \citep{pml2Book}.

 Lastly, we calculate the canonical vectors $(\widehat{a},\widehat{b})$ and correlation $\widehat{\rho}_c$ with the index sets of estimated associated variables $(\widehat{I}_X,\widehat{I}_Y)$ based on \eqref{eq:cvec} and \eqref{eq:ccor}. The computational complexity of the greedy algorithm is $O(p+q)$ when a single subgraph is present, while it increases with the number of subgraphs, reaching a worst-case complexity of $O((p+q)^2)$. 

%\tcr{(Ray: how is $c^\star$ chosen?)} \tblue{I tried to consider $C$ as unknown, but it would be better to put it known in this work. }

\begin{algorithm}[h]
\caption{: Greedy Algorithm for Graph Canonical Correlation Analysis}
\begin{algorithmic}[1]
            \State \textbf{Input:} joint data set ${\bf X}$ and ${\bf Y}$, threshold $\varepsilon \in (0,1)$, $\bm{\lambda} = \{\lambda_1,\dots,\lambda_{g_0}\}$ 
		\State \textbf{Output:} canonical vectors $(\widehat{a}$, $\widehat{b})$ and canonical correlation $\widehat{\rho}_c$
          \State Set the initial values $\mathbf{R}^{1,1} = |\mathbf{R}^\varepsilon|$, active set $J^{1,1} = (J_X^{1,1},J_Y^{1,1},E^{1,1}=\{e_{ij}:(i,j)\in J_X^{1,1}\otimes J_Y^{1,1}\})$ s.t $J_X^{1,1} =[p]$, $J_Y^{1,1} =[q]$
     \For{$g = 1,2,\dots,g_0$}
        \For{$c = 1,2,\dots,C$}
             \For{$t = 1,2, \dots, \text{r.size}(\mathbf{R}^{c,1})+\text{c.size}(\mathbf{R}^{c,1})-1$}
            \State Set $\tau_t=\text{argmin}_\tau(\text{r.means}(\mathbf{R}^{c,t})_\tau)$ and $\phi_t = \text{argmin}_{\phi}(\text{c.means}(\mathbf{R}^{c,t})_\phi)$
            \If {$\text{r.means}(\mathbf{R}^{c,t})_{\tau_t} > \text{c.means}(\mathbf{R}^{c,t})_{\phi_t}$ }
            \State $J_X^{c,t+1} \leftarrow J_X^{c,t}$, and $J_Y^{c,t+1} \leftarrow J_Y^{c,t}\backslash \{v_{\phi_t}^{c,t}\}$, $\mathbf{R}^{c,t+1} \leftarrow \mathbf{R}^{1,1}[J_X^{c,t+1},J_Y^{c,t+1}]$
            \Else {}
            \State $J_X^{c,t+1} \leftarrow J_X^{c,t}\backslash \{u_{\tau_t}^{c,t}\}$, and $J_Y^{c,t+1} \leftarrow J_Y^{c,t}$, $\mathbf{R}^{c,t+1} \leftarrow \mathbf{R}^{1,1}[J_X^{c,t+1},J_Y^{c,t+1}]$
            \EndIf
            \State $J^{c,t+1} \leftarrow (J_X^{c,t+1},J_Y^{c,t+1},E^{c,t+1}=\{e_{ij}:(i,j)\in J_X^{c,t+1}\otimes J_Y^{c,t+1}\})$
            \EndFor  
            \State Set $B_c = \text{argmax}_{J^{c,t}} f_{\lambda_g}(J^{c,t},\varepsilon)$, where $B_c = (U_c,V_c,E_c)$
            \State $J^{c+1,1}_X \leftarrow [p]\backslash \cup^{c}_{c'=1} U_{c'}$, $J^{c+1,1}_Y \leftarrow [q]\backslash\cup^{c}_{c'=1} V_{c'}$, $\mathbf{R}^{c+1,1} \leftarrow \mathbf{R}^{1,1}[J^{c+1,1}_X,J^{c+1,1}_Y]$
        \EndFor
        \State $B^{\lambda_g} \leftarrow \{B_c\}_{c=1}^{C}$
        \EndFor
        \State Find $g^\star = \text{argmax}_{g=1,2,\dots,g_0} D_{KL}(P_g||Q)$ and set $\widehat{B} \leftarrow B^{\lambda_{g^\star} }$, where $\widehat{B} = \{(\widehat{U}_c,\widehat{V}_c,\widehat{E}_c)\}_{c=1}^C$
        \State Set $\widehat{I}_X = \cup_{c=1}^{C} \widehat{U}_c$ and $\widehat{I}_Y = \cup_{c=1}^{C} \widehat{V}_c$
        \State Return the estimated canonical vectors $(\widehat{a},\widehat{b})$ and correlation $\widehat{\rho}_c$ calculated based on \eqref{eq:cvec} and \eqref{eq:ccor}
	\end{algorithmic}
\label{algo1}
\end{algorithm}

%for $(i,j)$ and $(i,j')$ not in the subgraph and a filtration  $\mathcal{F} = \sigma\{R_{ij},R_{i'j'}\}_{j'\in I_Y'}$, where $\mathcal{F} = \sigma\{R_{i'j},R_{i'j'}\}_{i'\in I_X'}$ and $I_X'$ is an arbitrary subset in $[p]\backslash I_X$ such that $i\notin I_X'$. 

\section{Theoretical Properties of gCCA}
\label{sec:3}

In this section, we outline the assumptions and present the theoretical results to show the minimum sample size condition for a high probability performance guarantee of the estimation procedure of gCCA.

\subsection{Assumptions}
We first introduce the assumptions underlying the theories in the gCCA estimation procedure. Our first assumption concerns the distributional properties of correlation coefficients in the large cross-correlation matrix ${\bf R}$. Specifically, we assume that the sample correlation coefficients, given a sample size of $n$, satisfy a concentration property, ensuring that they are sufficiently bounded with high probability \citep{boucheron2003concentration}. This assumption is formulated based on the concentration inequality for the sample correlation coefficient of two bivariate normal variables, as provided in Subsection \ref{ssec:ass1} of Supplementary Materials.

% We first introduce the assumptions required to analyze the greedy algorithm for gCCA. Our first assumption is a concentration property of sample correlation coefficients with sample size $n$ to ensure that sample correlation coefficients are concentrated sufficiently. This assumption is formulated in light of the concentration inequality for a sample correlation coefficient of two bivariate variables from normal distributions provided in Supplementary Materials.
\begin{assumption}[$n^{-1/2}$ concentration of sample correlation coefficients]
    Let $R_{ij}$ be the sample correlation coefficient of $X_i$ and $Y_j$ with size $n> 3$, which is generated from a bivariate distribution with a ground truth correlation coefficient $\rho_{ij} \in [-1+\delta,1-\delta]$ for a $0<\delta<1$. Then, for all $a>0$ and all $\rho_{ij} \in [-1+\delta,1-\delta]$, there are $s_1,~s_2>0$ such that
    \begin{eqnarray*}
        P(|R_{ij}-\rho_{ij}| > a) \leq s_1 \exp \left(-\frac{na^2}{2s_2^2}\right).
    \end{eqnarray*}
    \label{ass:sub}
\end{assumption}
%\tcr{Ray: does the ``approximate concentration inequality'' in \citep{salnikov2024concentration} imply the bound above? }
%\tcb{\cite{salnikov2024concentration} shows $ P(|R_{ij}-\rho_{ij}| > a) \leq 2\exp \left(-\frac{na^2}{2s^2}\right)$ approximately for normal distribution.}
%\tcr{Ray: what does ``approximately'' mean? In partcular, what does ``$\lesssim$'' precisely mean in that paper?  Does it mean the bound $ 2\exp \left(-\frac{na^2}{2s^2}\right)$ holds up to a constant (this is the usual understanding of $\lesssim$)? If this is the case, then $2$ in $2\exp \left(-\frac{na^2}{2s^2}\right)$ may not be meaningful.  I think it is particularly important to ensure that we do have an exponential bound  $c_1\exp(-n a^2 c_2)$  that does hold for all $n$, not just to have asymptotics since we claim to have finite-sample guarantee. } 
%\tcb{Thanks for pointing this out. $\lesssim$ in the paper means that the inequality works for a large $n$. But, in the paper, $s$ is set to be $1-\rho^2$. Because we say that there is $s$ satisfying the inequality for $n\geq 3$, I think that it is OK for a small $n$. In addition, the paper demonstrates that the inequality with $s=1-\rho^2$ is satisfied for $n=10$ via a simulation study. Even with the result of the simulation study, I think that we can set this assumption.}
%\tcr{(Ray: does the subGaussian assumption on the right implies already the variance bound on the left? $  Var(R_{ij})= \int_0^\infty 2a P(|R_{ij}-\rho_{ij}| > a)  da $).} \tblue{That is true. the second inequality implies the first one. I will remove the first one.}
Assumption \ref{ass:sub} is equivalent to the fact that there exist $s_3,s_4>0$ such that $E[e^{a(R_{ij}-\rho_{ij})}] \leq s_3 e^{s_4^2a^2/n}.
$ for all $a>0$. This equivalence is similarly constructed to the equivalent definitions of subGaussian random variables \citep{vershynin2018high}. This is shown in Subsection \ref{ssec:ass11} of Supplementary Materials. In the case with $s_1=2$ and $s_3=2$, we can say that $R_{ij}$ is $s_2/\sqrt{n}$-subGaussian. Second, we assume the independency of two correlation coefficients, if the indices of the two correlation coefficients are not in the subgraph.  %for the difference between two truncated absolute sample correlation coefficients given a piece of information of some known truncated absolute correlation coefficients, the index set of which is $I_Y'$. 
%\tcr{(Ray: would it be worth clarifying that $J_X^t$ and $J_Y^t $ refer to the first round of exclusion?)}
%\begin{assumption}
%    For given  $i_1 \notin I_X$, $i_2 \in [p]$, and $j_1 \notin I_Y$, we have 
%    \begin{eqnarray*}
%        E[e^{a (|R_{i_1j_1}^{\varepsilon}|-|R_{i_2j_1}^{\varepsilon}|)}|\mathcal{F}] \leq e^{s^2a^2/n },~\text{for~all}~a>0,
%    \end{eqnarray*}
%    where $\mathcal{F} = \sigma\{ \{|R_{i_1j'}^{\varepsilon}|,|R_{i_2j'}^{\varepsilon}|\}_{j\in I_Y'} \}$ for an arbitrary such that $I_Y' \cap I_Y =\emptyset$ and $j_1 \notin I_Y'$. Similarly, we have $E[e^{a (|R_{i_1j_1}^\varepsilon|-|R_{i_1j_2}^\varepsilon|)}|\mathcal{F}] \leq e^{s^2a^2/n }$ with the relevant quantities defined in the same way as above.\label{ass:ass2}
%\end{assumption} 

\begin{assumption}
    For $(i_1, j_1),~(i_2, j_1) \notin I_X \otimes I_Y$, 
    $R_{i_1j_1}$ and $R_{i_2j_1}$ are independent.  Similarly, for $(i_1, j_1),~(i_1, j_2) \notin I_X \otimes I_Y$, 
    $R_{i_1j_1}$ and $R_{i_1j_2}$ are independent.    \label{ass:ass2}
\end{assumption} 
Assumption \ref{ass:sub} provides a $n^{-1/2}$ concentration inequality for each correlation coefficient, while Assumption \ref{ass:ass2} provides a useful property of the joint distribution of two. The statement in Assumption \ref{ass:ass2} indeed holds for normal distributions, because a centered and standardized $d$-dimensional normal random vector has a uniform distribution in $\mathbb{S}^{d-1}$ and accordingly the angle (correlation coefficient) between two independent vectors does not provide any information about their individual directions (For further details, see \ref{ssec:ass2} in Supplementary Materials).

\subsection{Theoretical results}
\label{ssec:results}
In this subsection, we present the main theoretical results for the gCCA estimation procedure. The first result demonstrates that the greedy algorithm can achieve full recovery of the variables with nonzero entries in the canonical vectors. The second result establishes the square-root estimation accuracy of the canonical correlation estimates obtained via gCCA.

Without loss of generality, we consider that the correlation matrix has a network structure with a single biclique subgraph $B=(U_1,V_1,E_1)$, where the absolute values of correlations $|\rho_{ij}|$ for all $(i,j) \in U_1\otimes V_1$ have the values in $[\rho,w\rho]$ for some $0<\rho<1$ and $w\geq 1$. In addition, the correlation coefficients outside the subgraph are zero. For ease of presentation, we set $I_X=\{1,2,\dots,p_0\}$ and $I_Y=\{1,2,\dots,q_0\}$. Also, we consider a threshold $\varepsilon=\rho/2$. The following lemma provides bounds to ensure that the sample correlation coefficients, both within and outside the subgraph, lie within specific intervals based on the subGaussianity of sample correlation coefficients. The complete proofs of the following lemmas are provided in Supplementary Materials.

% In this subsection, we demonstrate two theoretical results. The first result demonstrates that the greedy algorithm can achieve full recovery of the associated variables associated with the canonical correlations. The second result establishes the square-root estimation accuracy of the canonical correlation obtained using the greedy algorithm.
% %we show that gCCA is guaranteed to identify all the related variables for associations between two random vectors $X\in\mathbb{R}^p$ and $Y\in\mathbb{R}^q$ with a high probability.
% We consider the setup where 
% the correlation matrix has only a biclique subgraph $B=(U_1,V_1,E_1)$, where the absolute values of correlations $|\rho_{ij}|$ for all $(i,j) \in U_1\otimes V_1$ have the values in $[\rho,w\rho]$ for some $0<\rho<1$ and $w\geq 1$. In addition, the correlation coefficients outside the subgraph are assumed to be 0. For ease of presentation, we set $I_X=\{1,2,\dots,p_0\}$ and $I_Y=\{1,2,\dots,q_0\}$. Also, we consider a threshold $\varepsilon=\rho/2$. The following lemma provides bounds to ensure that the sample correlation coefficients, both within and outside the subgraph, lie within specific intervals based on the subGaussianity of sample correlation coefficients.

\begin{lemma}
    Let $R_{ij}$ and $\rho_{ij}$ be the sample correlation coefficient with sample size $n$ and ground truth correlation of $X_i$ and $Y_j$, respectively. Then, for all $i \in [p]$ and $j \in [q]$, with probability at least $1-\delta$, we have
\begin{eqnarray*}
   |R_{ij}-\rho_{ij}| < \sqrt{\frac{2s_2^2}{n}\log \left(\frac{s_1pq}{\delta}\right)}.
\end{eqnarray*}
\label{lem:1}
\end{lemma}

The next lemma is an anti-concentration inequality for the difference of sums of absolute sample correlation coefficients between one in a subgraph and another outside it. This inequality makes the two sums (one in a subgraph, another outside it) distinct so that we can exclude rows or columns not in subgraphs.

\begin{lemma}
    For the subgraph $B$, if $n\geq (\eta/\rho)^2 (2s_2\log \left(s_1pq/\delta\right))$, we have $\sum_{j\in I_Y} (|R_{i_1j}^\varepsilon|  - |R_{i_2j}^\varepsilon|) > q_0\rho(\eta-2)/\eta$ for $i_1\in I_X$, $i_2\in I_X^c$, and $\eta>3$ with probability at least $1-\delta$. Similarly, $\sum_{i\in I_X} (|R_{ij_1}^\varepsilon|  - |R_{ij_2}^\varepsilon|) > p_0\rho(\eta-2)/\eta$ for $j_1\in I_Y$ and $j_2\in I_Y^c$ with probability at least $1-\delta$.   \label{lem:2}
\end{lemma}

The following lemma presents a concentration inequality for the differences of the column means of $|R_{ij}^\varepsilon|$ between two columns. This inequality ensures that the error terms for the sample correlation coefficients of the uncorrelated variables remain bounded over the consecutive exclusion process. As a result, the greedy algorithm can sequentially eliminate uncorrelated variables with high probability.

\begin{lemma}
    Let $h_1$ and $h_2$ be bijective functions from $\{1,2,\dots,q-q_0\}$ to $I_Y^c$ and from $\{1,2,\dots,p-p_0\}$ to $I_X^c$, respectively. with probability at least $1-\delta$, if $n > \frac{\eta^2 s_4^2}{\min(p_0,q_0)^2(\eta-2)^2\rho^2}\left( \max(p,q)(1+\log s_3)+ \log \frac{2\max(p,q)^2}{\delta}\right)^2$, we have
    $$\sum_{j=1}^t (|R_{i_1h_1(j)}^\varepsilon| - |R_{i_2h_1(j)}^\varepsilon|) \leq \frac{\min(p_0,q_0)(\eta-2)\rho}{\eta},$$
    for all $i_1 \notin  I_X$, $i_2\in [p]$, all $1\leq t \leq q-q_0$. Similarly, with probability $1-\delta$, we have 
    $$\sum_{i=1}^{t} (|R_{h_2(i)j_1}^\varepsilon| - |R_{h_2(i)j_2}^\varepsilon| ) \leq \frac{\min(p_0,q_0)(\eta-2)\rho}{\eta},$$
    for all $j_1\notin I_Y$, $j_2\in [q]$ and all $1\leq t \leq p-p_0$. 
    \label{lem:3}
\end{lemma}

Now, we are ready to prove the main result based on the three lemmas above. The following result demonstrates that gCCA can detect all the related and irrelevant variables for associations of two high-dimensional variables with a high probability when the minimum sample condition is satisfied.

\begin{theorem}
    For $\delta \in (0,1)$, with probability at least $1-\delta$, if 
\begin{eqnarray*}
    n> (1/\rho^2) \left(2\sqrt{2s_2^2\log \left(2s_1pq/\delta\right)} + \frac{s_4}{2\min(p_0,q_0)}\left(\max(p,q)(1+\log s_3)+ \log \frac{4\max(p,q)^2}{\delta} \right)\right)^2,
\end{eqnarray*} there is a set of $\lambda$ in $(0,1)$ to guarantee that the greedy algorithm for gCCA has
\begin{eqnarray*}
    \mathbb{P}(I_X = \widehat{I}_X~\text{and}~I_Y = \widehat{I}_Y) \geq 1-\delta,
\end{eqnarray*}
where $\widehat{I}_X$ and $\widehat{I}_Y$ are the sets of associated variables estimated by the greedy algorithm. 
\label{thm:1}
\end{theorem}
\textbf{proof sketch: } First, based on Lemma \ref{lem:1}, \ref{lem:2}, and \ref{lem:3}, we bound the magnitude of errors created by samples so that the associated variables are distinguishable. Then, we show that the greedy algorithm sequentially excludes uncorrelated variables from our active sets until only the associated variables remain. Next, we demonstrate that there exists a set of $\lambda$ such that our objective function is maximized at the time when only and all uncorrelated variables are excluded. Lastly, we find the value $\eta$ to minimize the sample size suggested in the lemmas.
\begin{proof}

To streamline the presentation, we define two sub-timelines $t_X$ and $t_Y$ corresponding to the rows and columns, respectively. $t_X-1$ and $t_Y-1$ are the numbers of rows and columns excluded from the initial active matrix, which is given by the truncated absolute correlation matrix $\mathbf{R}^1=|\mathbf{R}^\varepsilon|$. Here, the overall time $t$ is related to the sub-timelines by the equation $t = t_X + t_Y - 1$. When a row is excluded, we update the row sub-timeline $t_X \leftarrow t_X + 1$. Otherwise, we have $t_Y \leftarrow t_Y + 1$. 

First, we show that the greedy algorithm excludes all rows in $I_X^c$ and columns in $I_Y^c$ before excluding ones in $I_X$ or $I_Y$. Let $i_{t_X}$ and $j_{t_Y}$ denote the row and column excluded at time $t_X$ and $t_Y$ in the sub-timelines, respectively. Assume that $\{i_{t_X}\}_{t_X \in [p] }$ and $\{j_{t_Y}\}_{t_Y\in [q]}$ are the sequences of row and column exclusion. It suffices to show that the sets of the first $p-p_0$ excluded rows and $q-q_0$ columns are identical to $I_X^c$ and $I_Y^c$, which means $\{i_{t_X}\}_{t_X\in[p-p_0]} = I_X^c$ and $\{j_{t_Y}\}_{t_Y\in[q-q_0]} = I_Y^c$. Let $\{i_{t_X}'\}_{t_X \in [p - p_0]}$ and $\{j_{t_Y}'\}_{t_Y \in [q - q_0]}$ be reversely enumerated subsequences of $\{i_{t_X}\}_{t_X \in [p]}$ and $\{j_{t_Y}\}_{t_Y \in [q]}$, respectively, where $i_{t_X}' \in I_X^c$ for all $t_X$ and $j_{t_Y}' \in I_Y^c$ for all $t_Y$. Note that 
\begin{eqnarray}
    \sum_{t_Y = 1}^t (|R_{i'j'_{t_Y}}^\varepsilon |-|R_{ij'_{t_Y}}^\varepsilon |) \leq  \frac{\min(p_0,q_0)(\eta-2)\rho}{\eta}\label{eq:reverse}
\end{eqnarray}
for all $i \in I_X$, $i'\in I_X^c$ and $t \in [q-q_0]$ by Lemma \ref{lem:3} with probability $1-\delta$ given that the minimum sample size condition is met.

Let $i_{t_X'} \in I_X$ be the first excluded row in $I_X$ before a column is excluded from $I_Y$. Suppose that there is $t_X''$ such that $t_X'<t_X''$ and $i_{t_X''}\in I_X^c$. At the exclusion of row $i_{t_X'}$, assume that the active columns at sub-timeline $t_Y'$ are $J_Y^{t_Y'} =\{j_{t_Y'+1},\dots,j_q\}$. Then,
as we exclude row $i_{t_X'}$, we have $\sum_{j\in J_Y^{t'_Y}} |R_{i_{t_X'}j}^\varepsilon| < \sum_{j\in J_Y^{t'_Y}} |R_{i_{t_X''}j}^\varepsilon|$. But, this contradicts the fact $\sum_{j\in I_Y} |R_{i_{t_X'}j}^\varepsilon| + \sum_{j\in J_Y^{t_Y'}\backslash I_Y} |R_{i_{t_X'}j}^\varepsilon| > \sum_{j\in J_Y^{t_Y'}} |R_{i_{t_X''}j}^\varepsilon| + \sum_{j\in J_Y^{t_Y'}\backslash I_Y}|R_{i_{t_X''}j}^\varepsilon|$, as $\sum_{j\in I_Y} |R_{i_{t_X'}j}^\varepsilon| > \frac{q_0(\eta-1)}{\eta}\rho$ and $\sum_{j\in I_Y} |R_{i_{t_X'}'j}^\varepsilon| < \frac{q_0}{\eta}\rho$ by Lemma \ref{lem:2}, and $\sum_{j\in J_Y^{t_Y'}\backslash I_Y} (|R_{i_{t_X''}j}^\varepsilon| - |R_{i_{t_X'}j}^\varepsilon|) <  \frac{\min(p_0,q_0)(\eta-2)\rho}{\eta}$ by \eqref{eq:reverse}. We can get the same contradiction when we assume that $j_{t_Y'} \in I_Y$ for $t_Y' \in [q-q_0]$ be the first excluded column in $I_Y$ before a row is excluded from $I_X$. Therefore, the greedy algorithm excludes all the rows in $I_X^c$ first and then ones from $I_X$. Similarly, after all the columns in $I_Y^c$ are excluded, ones in $I_Y$ are excluded.

Now, it suffices to show that $i \in I_X$ is not excluded before $j' \in I_Y^c$ is excluded. Suppose all $i'\in I_X^c$ are excluded and some $j' \in I_Y^c$ are left in the active set at time $t_X =\tau_X$ and $t_Y =\tau_Y$. Then, the active sets are $J_X^{\tau_X} = I_X$ and $J_Y^{\tau_Y} = I_Y \cup I_Y'$ for a non-empty set $I_Y' \subset I_Y^c$. We consider the row sum of row $i\in I_X$, $\sum_{j\in I_Y} |R_{ij}^\varepsilon| + \sum_{j \in I_Y'} |R_{ij}^\varepsilon|$, and column sum of column $j'\in I_Y^c$, $\sum_{i\in I_X} |R_{ij'}^\varepsilon|$ with respect to the active sets $J_X^{\tau_X}$ and $J_Y^{\tau_Y}$. Because $\sum_{i\in I_X} |R_{ij'}^\varepsilon|<\min(p_0,q_0)\rho/\eta$ for $j' \in I_Y^c$; $\sum_{j\in I_Y} |R_{ij}^\varepsilon| \geq \min(p_0,q_0)\rho(\eta-1)/\eta$, $\sum_{j \in I_Y'} |R_{ij}^\varepsilon| < \min(p_0,q_0)\rho/\eta$ for $i \in I_X$; and $\eta>3$,
we have $\sum_{j\in I_Y} |R_{ij}^\varepsilon| + \sum_{j \in I_Y'} |R_{ij}^\varepsilon| > \sum_{i\in I_X} |R_{ij'}^\varepsilon|$. Thus, we exclude all $j'\in I_Y^c$ before we start excluding $i\in I_X$. Similarly, we exclude $i'\in I_X^c$ before we start excluding $j\in I_Y$.

Next, we show that there is a set of $\lambda$ such that the objective function $f_\lambda$ is maximized at the time when only and all the associated rows and columns remain in our active set. First, we compare two cases: (i) only all rows $I_X$ and columns in $I_Y$ remain in our active set, (ii) $r_1$ rows in $I_X^c$ and $c_1$ columns $I_Y^c$, the index sets of which are denoted by $I_X''$ and $I_Y''$,  remain in our active set including all $i \in I_X$ and $j\in I_Y$ at time $t$. Then, the values of the objective functions for case (i) and (ii) are as follows:
\begin{eqnarray*}
f_\lambda(B_{case_1},\varepsilon)=\frac{\sum_{i\in I_X,j\in I_Y} |R_{ij}^\varepsilon|}{(p_0 q_0)^\lambda}
\end{eqnarray*}
and
\begin{flalign*}
&f_\lambda(B_{case_2},\varepsilon) \\
&=\frac{\sum_{i\in I_X,j\in I_Y} |R_{ij}^\varepsilon| + \sum_{i\in I_X,j\in I_Y''} |R_{ij}^\varepsilon| +\sum_{i\in I_X'',j\in I_Y} |R_{ij}^\varepsilon| + \sum_{i\in I_X'',j\in I_Y''} |R_{ij}^\varepsilon|}{( (p_0+r_1)(q_0+c_1) )^\lambda}
\end{flalign*}

Applying $|R_{ij}^\varepsilon| > ((\eta-1)/\eta)\rho$ for $(i,j) \in I_X\otimes I_Y$ and $|R_{ij}^\varepsilon| < (1/\eta)\rho$ otherwise, we derive the following inequality:
\begin{eqnarray}
\lambda > \log \left(\frac{\eta-2}{\eta-1} + \frac{1}{\eta-1}\frac{(p_0+r_1)(q_0+c_1)}{p_0q_0}\right)/\log \left(\frac{(p_0+r_1)(q_0+c_1)}{p_0q_0}\right)    \label{eq:lambda1}
\end{eqnarray}

We consider case (iii) such that only some $r_2$ rows in $I_X$ and $c_2$ columns in $I_Y$ remain in our active sets, which are denoted by $J_X'$ and $J_Y'$. Then, we have
    
\begin{eqnarray*}
    f_\lambda(B_{case_3},\varepsilon)= \frac{\sum_{i\in J_X',j\in J_Y'} |R_{ij}^\varepsilon|}{( r_2c_2 )^\lambda}.
\end{eqnarray*}

Now, we find the range of $\lambda$ such that the objective function of case (i) is greater than that of case (iii). By applying the upper and lower bounds of $|R_{ij}^\varepsilon| \leq ((\eta+w)/\eta)\rho$ for $(i,j) \in J_X'\otimes J_Y'$ and $|R_{ij}^\varepsilon| \geq ((\eta-1)/\eta)\rho$ for $(i,j) \in (I_X\otimes I_Y)\backslash (J_X'\otimes J_Y')$, respectively, for the inequality $\frac{\sum_{i\in I_X,j\in I_Y} |R_{ij}^\varepsilon|}{(p_0 q_0)^\lambda} \geq  \frac{\sum_{i\in J_X',j\in J_Y'} |R_{ij}^\varepsilon|}{( r_2c_2)^\lambda}$, we have
\begin{eqnarray}
\lambda < \log \left(\frac{w+1}{\eta+w} + \frac{\eta-1}{\eta+w}\frac{p_0q_0}{r_2c_2}\right)/\log \left(\frac{p_0q_0}{r_2c_2}\right).\label{eq:lambda2}
\end{eqnarray}
Putting \eqref{eq:lambda1} and \eqref{eq:lambda2} together, we have
\begin{eqnarray*}
    0<\frac{\log \left(\frac{\eta-2}{\eta-1} + \frac{1}{\eta-1}\frac{(p_0+r_1)(q_0+c_1)}{p_0q_0}\right)}{\log \left(\frac{(p_0+r_1)(q_0+c_1)}{p_0q_0}\right)} <\lambda< \frac{\log \left(\frac{w+1}{\eta+w} + \frac{\eta-1}{\eta+w}\frac{p_0q_0}{r_2c_2}\right)}{\log \left(\frac{p_0q_0}{r_2c_2}\right)} < 1.
\end{eqnarray*}
To satisfy the above inequality for all $1\leq r_1\leq p-p_0$, $1\leq c_1< q-q_0$, $0<r_2<p_0$, and $0<c_2<q_0$, we need sufficiently large $\eta$. If we apply the ranges of $r_1,~c_1,~r_2$ and $c_2$, we can show that the inequality can be written as
\begin{eqnarray*}
   \log\left(\frac{\eta-2}{\eta-1} +\frac{1}{\eta-1} a_1\right)/\log a_1 < \log\left( \frac{w+1}{\eta+1} +\frac{\eta-1}{\eta+w}a_2 \right)/\log a_2,
\end{eqnarray*}
where $a_1 = \min(\frac{p_1+1}{p_0},\frac{q_0+1}{q_0})>1$ and $a_2 = \min(\frac{p_0}{p_0-1},\frac{q_0}{q_0-1})>1$. As $a_1<a_2$ and the term on the RHS above is increasing in $a_2 > 1$, we can replace $a_2$ with $a_1$. Then, we get $\log\left(\frac{\eta-2}{\eta-1} +\frac{1}{\eta-1} a_1\right) < \log\left( \frac{w+1}{\eta+w} +\frac{\eta-1}{\eta+w}a_1\right)$. By simple calculation using $a_1>1$, we have $ \eta^2 - 3\eta +(1-w) > 0$. Using the quadratic formula, we have 
\begin{eqnarray}
\eta > (3+\sqrt{9-4(1-w)})/2\geq 3,  \label{eq:quad}
\end{eqnarray}
where $w\geq 1$. Thus, we show that there exists $\lambda$ satisfying that the objective function is maximized at the time when only all rows in $I_X$ and columns $I_Y$ are in our active set.

Lastly, we choose $\eta$ to minimize the minimum sample sizes in Lemma \ref{lem:2} and \ref{lem:3}. Now, the minimum sample size required for the result above is represented with respect to $\eta$ as follows:
\begin{eqnarray*}
    n> \max\left( 2s_2^2\left(\frac{\eta}{\rho}\right)^2 \log \left(\frac{s_1pq}{\delta}\right),\frac{\eta^2 s_4^2 \left( \max(p,q)(1+\log s_3)+ \log \frac{2\max(p,q)^2}{\delta}\right)^2}{4\min(p_0,q_0)^2(\eta-2)^2\rho^2}\right).
\end{eqnarray*}
As the first and second terms above are increasing and decreasing in $\eta$, respectively, we find the value $\eta^\star$ to make the two terms equal. Thus, we get
\begin{eqnarray*}
    \eta^\star = 2 + \frac{\frac{s_4}{2\min(p_0,q_0)}\left(\max(p,q)(1+\log s_3)+ \log \frac{2\max(p,q)^2}{\delta} \right)}{\sqrt{2s_2^2\log \left(s_1pq/\delta\right)}}.
\end{eqnarray*}
When $\eta^\star$ satisfies the inequality \eqref{eq:quad}, by plugging $\eta^\star$ into the sample size, we get
\begin{eqnarray*}
    n> (1/\rho^2) \left(2\sqrt{2s_2^2\log \left(s_1pq/\delta\right)} + \frac{s_4}{2\min(p_0,q_0)}\left(\max(p,q)(1+\log s_3)+ \log \frac{2\max(p,q)^2}{\delta} \right)\right)^2,
\end{eqnarray*}
with probability $1-2\delta$, which is a combined probability of the applications of Lemma \ref{lem:2} and \ref{lem:3}. By plugging $\delta/2$ into $\delta$, we get the desired result.
In the most practical scenarios, $\eta^\star$ satisfies the inequality \eqref{eq:quad}. Otherwise, we set $\eta^{\star}_2 = (3+\sqrt{9-4(1-w)})/2$. Then, we get $n> 2s_2^2\left(\frac{\eta^\star_2}{\rho}\right)^2 \log \left(\frac{s_1pq}{\delta}\right)$. \hfill$\square$
\end{proof}
The above results provide the greedy algorithm \citep{charikar2000greedy} with a minimum sample size condition ($O(\max(p,q)^2/\min(p_0,q_0)^2$)) for the full recovery property with high probability under the conditions described in the previous sub-subsection. The next theorem provides the square-root estimation accuracy of the canonical correlation of the greedy algorithm. 

\begin{theorem}
    For $\delta \in (0,1)$, if 
\begin{eqnarray*}
    n> (1/\rho^2) \left(2\sqrt{2s_2^2\log \left(2s_1pq/\delta\right)} + \frac{s_4}{2\min(p_0,q_0)}\left(\max(p,q)(1+\log s_3)+ \log \frac{4\max(p,q)^2}{\delta} \right)\right)^2,
\end{eqnarray*}
the estimated canonical correlation $\widehat{\rho}_c$ calculated based on \eqref{eq:ccor} by the greedy algorithm has a square-root estimation consistency with respect to the sample size $n$ such that
    \begin{eqnarray*}
        \mathbb{P}\left(|\widehat{\rho}_c - \rho_c| < \sqrt{\frac{2s_2^2p_0q_0}{n}\log \frac{s_1pq}{\delta}}  \right)  \geq 1- \delta.
    \end{eqnarray*}
    \label{thm:2}
\end{theorem}
\begin{proof}
First, note that the greedy algorithm identifies the associated variables with probability at least $1-\delta$, provided the minimum sample condition in Theorem \ref{thm:1} is satisfied. Accordingly, we suppose that $I_X =\widehat{I}_X$ and $I_Y =\widehat{I}_Y$ with probability at least $1-\delta$. Now, we calculate $\widehat{\rho}_c$ based on \eqref{eq:ccor}. By Lemma \ref{lem:1}, we have
    \begin{eqnarray*}
        |R_{ij} - \rho_{ij}| < \sqrt{\frac{2s_2^2}{n}\log \frac{s_1pq}{\delta}} 
    \end{eqnarray*}
for all $i\in[p]$ and $j\in [q]$ with probability at least $1-\delta$. Thus, we have
\begin{eqnarray*}
    \|\Sigma_{X_0Y_0} - {\bf X}^\top_0 {\bf Y}_0\|_F = \sqrt{\sum_{i\in[p_0],j\in[q_0]} (R_{ij}-\rho_{ij})^2} \leq \sqrt{\frac{2s_2^2p_0 q_0}{n} \log \frac{s_1pq}{\delta}},
\end{eqnarray*}
where $\|M\|_F$ is the Frobenius norm of a matrix $M$. Let $\rho_c = \| \Sigma_{XY} \|_2=\| \Sigma_{X_0Y_0} \|_2$, where $\text{Cov}(X,Y) =\Sigma_{XY}$ and $\text{Cov}(X_0,Y_0) =\Sigma_{X_0Y_0}$. As $\|\Sigma_{X_0Y_0}\|_2 - \|{\bf X}^\top_0 {\bf Y}_0\|_2 \leq \|\Sigma_{X_0Y_0} - {\bf X}^\top_0 {\bf Y}_0\|_2$, $\|{\bf X}^\top_0 {\bf Y}_0\|_2 - \|\Sigma_{X_0Y_0}\|_2 \leq \|\Sigma_{X_0Y_0} - {\bf X}^\top_0 {\bf Y}_0\|_2$, and $\|\Sigma_{X_0Y_0} - {\bf X}^\top_0 {\bf Y}_0\|_2 \leq \|\Sigma_{X_0Y_0} - {\bf X}^\top_0 {\bf Y}_0\|_F$, we have $$|\rho_c - \widehat{\rho}_c| \leq \|\Sigma_{X_0Y_0} - {\bf X}^\top_0 {\bf Y}_0\|_F \leq \sqrt{\frac{2s_2^2p_0 q_0}{n} \log \frac{s_1pq}{\delta}} ,$$ with probability at least $1-\delta$. \hfill$\square$
\end{proof}

%To the best of our knowledge, non-asymptotic analysis of square-root estimation consistency in CCA has not been explored, although some asymptotic analyses have been conducted in this area \citep{anderson1999asymptotic}.

\section{Simulation}
\label{sec:4}

We numerically evaluate the performance of the proposed gCCA approach and benchmark it with existing CCA methods. We first simulate 500 samples ($n=500$) of $X$ and $Y$ with dimensions $p= 1000$ and $q= 1500$ based on
\begin{eqnarray*}
\begin{pmatrix}
    X     \\
    Y     \\
\end{pmatrix}\sim \mathcal{N}(\mathbf{0}_{p+q},\Sigma),~\Sigma=\begin{pmatrix}
    \Sigma_X & \Sigma_{XY}     \\
     \Sigma_{XY}^\top &  \Sigma_Y     \\
\end{pmatrix}
\end{eqnarray*}
where $\text{diag}(\Sigma) = \mathbf{1}_{p+q}$ and a bipartite graph for $\Sigma_{XY}$ has a latent biclique subgraph $B_1 = (U_1,V_1,E_1)$, where $I_X=U_1$ and $I_Y=V_1$ due to $C=1$. $X_i$ and $Y_j$ have a nonzero correlation $\rho_{ij} \neq 0$, if for $i\in U_1$ and $j\in V_1$; otherwise, $\rho_{ij}=0$. We perform the experiments for four different setups of the combinations of two subgraph sizes, $(|I_X|,|I_Y|)=(20,30)$ and $(|I_X|,|I_Y|)=(30,40)$ and two sets of correlations, $\rho_{ij} \in [0.2,0.3]$ and $\rho_{ij} \in [0.3,0.4]$. For each setup, we choose the best tuning parameter value from 0.5 to 0.9, with increments of 0.05 based on the KL divergence. To ensure a fair comparison with sCCA, we optimize the performance of sCCA by selecting tuning parameters via cross-validation. For each setting, we generate 100 data sets to evaluate the performance of gCCA and benchmark with sCCA \citep{witten2009extensions,witten2009penalized} using the criteria of sensitivity and specificity for estimates $\widehat{I}_X$. We also assess the proportion of both sensitivity and specificity equal to 1, which corresponds to the case of $\widehat{I}_X=I_X$ and $\widehat{I}_Y=I_Y$. In addition, we evaluate the bias, variance, and mean squared error (MSE) of estimates of canonical correlation $\widehat{\rho}_c$.

First, we assess the sensitivity and specificity of gCCA and sCCA for the four settings. In the context of classification, \textit{sensitivity}  measures the proportion of correlated $X$ and $Y$ pairs that are correctly identified by the model. It is defined as:
\begin{eqnarray*}
\text{Sensitivity} =  \frac{ |\widehat{I}_X\cap I_X| +|\widehat{I}_Y\cap I_Y|}{\underbrace{|\widehat{I}_X\cap I_X| +|\widehat{I}_Y\cap I_Y|}_{\text{number of true positives}} + \underbrace{|\widehat{I}_X^c\cap I_X| +|\widehat{I}_Y^c\cap I_Y|}_{\text{number of false negatives}} }.    
\end{eqnarray*}
  %A sensitivity of 1 indicates that all true positives (true associated variables in this work) are correctly detected.
On the other hand, \textit{specificity}  measures the proportion of true uncorrelated $X$ and $Y$ pairs that are correctly identified by the model. It is defined as:
\begin{eqnarray*}
\text{Specificity} = \frac{ |\widehat{I}_X^c\cap I_X^c| +|\widehat{I}_Y^c\cap I_Y^c|}{\underbrace{|\widehat{I}_X^c\cap I_X^c| +|\widehat{I}_Y^c\cap I_Y^c|}_{\text{number of true negatives}} + \underbrace{|\widehat{I}_X\cap I_X^c| +|\widehat{I}_Y\cap I_Y^c|}_{\text{number of false positives}}}.    
\end{eqnarray*}
%A specificity of 1 indicates that all true negatives are correctly identified. Thus, both sensitivity and specificity being 1 represent that all true positives and negatives are correctly identified. 
Table \ref{table:2} validates the performance of gCCA as presented in Theorem \ref{thm:1} from the previous section. gCCA demonstrates high sensitivity and specificity, together with a high proportion of both sensitivity and specificity equal to 1 for all four setups. In comparison, sCCA exhibits inconsistent performance: its high sensitivity and low specificity show that an excessive number of variables are identified as true positives in the first two setups, whereas in the remaining setups, it underestimates the number of true positives, leading to low sensitivity.

Next, we evaluate the bias, variance, and MSE of the canonical correlations estimated by gCCA and sCCA. The \textit{bias} of an estimator $\hat{\theta}$ for a parameter $\theta$ is the difference between the expected value of the estimator and the true value of the parameter: $\text{Bias}(\hat{\theta}) = \mathbb{E}[\hat{\theta}] - \theta$,  while the \textit{variance} of an estimator $\hat{\theta}$ quantifies how much $\hat{\theta}$ varies across different samples with the following definition: $\text{Var}(\hat{\theta}) = \mathbb{E}[(\hat{\theta} - \mathbb{E}[\hat{\theta}])^2]$. Lastly, the MSE combines both bias and variance. It is the expected squared difference between the estimator $\hat{\theta}$ and the true parameter $\theta$: $\text{MSE}(\hat{\theta}) = \mathbb{E}[(\hat{\theta} - \theta)^2]$. This can also be decomposed into bias and variance as: $\text{MSE}(\hat{\theta}) = \text{Bias}(\hat{\theta})^2 + \text{Var}(\hat{\theta})$. The MSE represents the total error of the estimator, taking into account both systematic error (bias) and random error (variance). 

Table \ref{table:3} demonstrates the bias, variance, and MSE of estimated canonical correlation values of gCCA and sCCA. gCCA consistently demonstrates more stable performance as compared to sCCA across all the setups. Considering the ratio of the MSE of gCCA to that of sCCA, the difference in performance is greater when a subgraph is small and its signal is weaker.

\begin{table}[]
\caption{Percentages of sensitivity, specificity and both being 1 of gCCA and sCCA for four different setups. The numbers in the parenthesis are the averages of sensitivity and specificity. The number in the parenthesis on every third line is the average of the geometric mean of sensitivity and specificity.}
\begin{tabular}{llll}\hline
         &                      & gCCA             & sCCA             \\\hline
                                                       & \% Sensitivity=1          & 90\% (0.998)        & $\mathbf{100\% (1)}$ \\ 
        $(|I_X|,|I_Y|) = (20,30)$, $\rho \in [0.2,0.3]$& \% Specificity=1          &  $\mathbf{100\% (1.000)}$ & 0\% (0.419) \\
                                                       & \% Both=1                &  $\mathbf{90\% (0.999)}$ & 0\% (0.647)  \\\hline
                                                       & \% Sensitivity=1         & 98\% (1.000)        & $\mathbf{100\% (1)}$ \\
        $(|I_X|,|I_Y|) = (20,30)$, $\rho\in [0.3,0.4]$ & \% Specificity=1         & $\mathbf{100\% (1.000)}$ & 0\% (0.419) \\
                                                       & \% Both=1                & $\mathbf{98\% (1.000)}$ & 0\% (0.647)  \\\hline
                                                       & \% Sensitivity=1          & $\mathbf{99\% (1.000)}$ & 53\% (0.933) \\
        $(|I_X|,|I_Y|) = (30,40)$, $\rho\in [0.2,0.3]$ & \% Specificity=1          & $\mathbf{100\% (1)}$ & 0\% (0.414) \\
                                                       & \% Both=1                 & $\mathbf{99\% (1.000)}$ & 0\% (0.622)     \\\hline
                                                       & \% Sensitivity=1          &  $\mathbf{99\% (1.000)}$        & 54\% (0.934)   \\
        $(|I_X|,|I_Y|) = (30,40)$, $\rho\in [0.3,0.4]$ & \% Specificity=1          &  $\mathbf{100\% (1)}$        & 0\% (0.416) \\
                                                       & \% Both=1                 &  $\mathbf{99\% (1.000)}$        & 0\% (0.623) \\\hline\\
\end{tabular}
\label{table:2}
\end{table}

\begin{table}[]
\caption{Bias, variance, and MSE of the estimated canonical correlation by gCCA and sCCA. }
\begin{tabular}{llll}\hline
                                                &                 & gCCA                           & sCCA                   \\\hline
                                                & $\text{Bias}^2$ & $\mathbf{7.645\times 10^{-7}}$ & $8.413\times 10^{-5}$  \\
$(|I_X|,|I_Y|) = (20,30)$, $\rho \in (0.2,0.3)$ & Variance        & $\mathbf{6.860\times 10^{-5}}$ & $7.6023\times 10^{-5}$ \\
                                                & MSE             & $\mathbf{6.936\times 10^{-5}}$ & $1.602\times 10^{-4}$  \\\hline
                                                & $\text{Bias}^2$ & $\mathbf{4.949\times 10^{-9}}$ & $1.573\times 10^{-5}$  \\
$(|I_X|,|I_Y|) = (20,30)$, $\rho\in (0.3,0.4)$  & Variance        & $\mathbf{3.412\times 10^{-5}}$ & $4.397\times 10^{-5}$  \\
                                                & MSE             & $\mathbf{3.413\times 10^{-5}}$ & $5.970\times 10^{-5}$  \\\hline
                                                & $\text{Bias}^2$ & $4.482\times 10^{-6}$ & $\mathbf{1.929\times 10^{-6}}$  \\
$(|I_X|,|I_Y|) = (30,40)$, $\rho\in (0.2,0.3)$  & Variance        & $\mathbf{3.532\times 10^{-5}}$ & $5.695\times 10^{-5}$  \\
                                                & MSE             & $\mathbf{3.980\times 10^{-5}}$ & $5.888\times 10^{-5}$  \\\hline
                                                & $\text{Bias}^2$ & $\mathbf{1.220\times 10^{-7}}$ & $3.287\times 10^{-7}$  \\
$(|I_X|,|I_Y|) = (30,40)$, $\rho\in (0.3,0.4)$  & Variance        & $\mathbf{1.641\times 10^{-5}}$ & $3.005\times 10^{-5}$  \\
                                                & MSE             & $\mathbf{1.653\times 10^{-5}}$ & $3.038\times 10^{-5}$ \\\hline\\
\end{tabular}
\label{table:3}
\end{table}

\section{Multi-omics Data Analysis}
\label{sec:5}
In this section, we apply gCCA to investigate the regulation of DNA methylation on gene expression in participants with Glioblastoma Multiforme (GBM) based on a data set from TCGA consortium \citep{tomczak2014cancer}. Alterations in DNA methylation within promoter regions have been widely documented in GBM, with such changes being associated with patient survival outcomes \citep{martinez2009microarray,giordano2014cancer}. Accordingly, identifying gene-specific methylation regulators in GBM is crucial for understanding the disease mechanisms and identifying potential therapeutic targets. For this study, we obtained DNA methylation data (measured using the HM27K array, covering around 27,000 CpG sites with zero-centered beta values) and gene expression data (measured by RNA-seq in RPKM) for the TCGA-GBM cohort from LinkedOmics. The data includes 278 samples of DNA methylation ($X$) and gene expression ($Y$) of the GBM cohort from The Cancer Genome Atlas (TCGA) database. The numbers of variables of $X$ and $Y$ are 6427 and 8196, respectively. 

In this study, our goal is to systematically investigate the regulatory effects of DNA methylation on gene expressions identifying i) which sets of DNA methylation variables are related to which sets of genes; and ii) measure the positive or negative correlations between them.  We applied gCCA to perform the analysis.

\begin{figure}[h]
\centering
\includegraphics[width=1\textwidth]{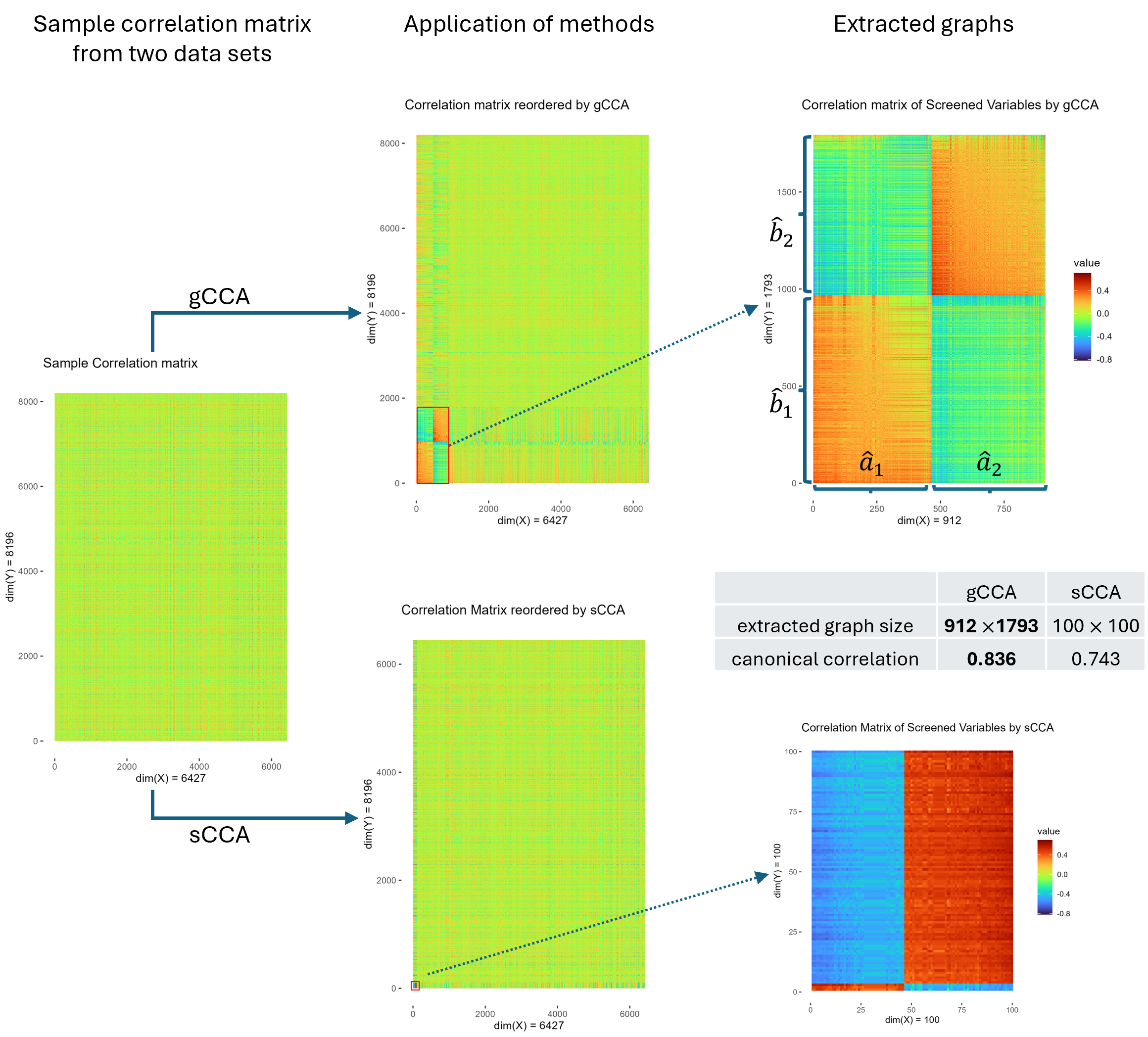}
\caption{Heatmaps of sample correlation matrices in the realdata analysis. The leftmost one is the sample correlation matrix. The two middle ones are the reordered correlation matrices by gCCA (top) and sCCA (bottom). The two figures on the rightmost are the extracted subgraphs from the TCGA-GBM data set by gCCA (top) and sCCA (bottom), respectively. The subgraphs extracted by gCCA and sCCA are of sizes 912 by 1793 and 100 by 100, respectively, with canonical correlations of 0.836 for gCCA and 0.743 for sCCA. }
\label{fig:3}
\end{figure}

We implement the greedy algorithm by objectively selecting 0.65 as the optimal tuning parameter based on the KL divergence. Figure \ref{fig:3} showcases a subgraph, which is extracted using the greedy algorithm for gCCA, with dimensions of 912 by 1793, organized into four distinct blocks showing positive and negative correlations between blocks with an overall canonical correlation of 0.836. In the plot of gCCA, the block, denoted by $(\hat{a}_1,\hat{b}_1)$ in the top right in Figure \ref{fig:3}, in the subgraph has strong associations 
%(Canonical correlation=0.573) 
with enzyme activities, particularly catalytic and kinase functions. This block contains methylation-gene pairs, including the pair cg05109049 with neurofibromatosis type I (NF1), and the pair cg10972821 with a kinase anchor protein 1 (AKAP1). NF1 functions as a GTPase-activating protein for RAS, a key driver of brain cancer with nerve glioma formation. AKAP1, a member of the A-kinase anchor protein family, is involved in binding to the regulatory subunit of protein kinase A in the cAMP-dependent signal pathway such as the mTOR pathway. High-level AKAP1 expression has been reported to activate the mTOR pathway, promoting glioblastoma growth.  

Additionally, we find the block, 
%(Canonical correlation=0.661)
denoted by $(\hat{a}_2,\hat{b}_2)$, which is strongly associated with the immune response, involving methylation-gene pairs such as cg21109025 paired with CCL2 (a member of CC chemokine family), cg17774418 paired with LY86 (lymphocyte antigen 86)  and cg21019522 paired with SLC22A18 (solute carrier family 22 member 18). These genes play crucial roles in recruiting immune cells to shape the tumor immune microenvironment (TIME). Methylation in these genes may directly influence their expression within TIME, potentially enhancing immune cytotoxicity while reducing immunosuppression mechanisms. Lastly, the blocks with negative correlations, denoted by $(\hat{a}_1,\hat{b}_2)$ and $(\hat{a}_2,\hat{b}_1)$, demonstrate that the two sets of genes, which are represented by $\hat{b}_1$ and $\hat{b}_2$, are associated with the methylations $(\hat{a}_1,\hat{a}_2)$ in the opposite way.

The subgraph extracted by sCCA with size 100 by 100 consists of variables with stronger associations than the average of those captured by gCCA. This subgraph has a canonical correlation of 0.743, which is lower than that of gCCA (0.836), due to its smaller size. This demonstrates that sCCA with the $\ell_1$ shrinkage can miss some relatively weaker signals of association between two high-dimensional variables, while the strongest signals are captured.

\section{Discussion}
\label{sec:6}

We have developed a new graph-based canonical correlation analysis tool - gCCA to decipher the systematic correlations between two sets of high-dimensional variables. Compared to traditional CCA methods, gCCA seeks not only to maximize the correlation between two canonical vectors but also to identify the latent patterns of correlated sets of variables taking the concept of a bipartite graph into account. gCCA can better differentiate the positively and negatively correlated variable sets, and yield canonical correlations to better assess the associations. The signs of canonical correlations are important for many applications like multi-omics data analysis, to reveal whether a set of variables positively/negatively affects the other set of variables. 

We provide a computationally efficient solution to implement gCCA with the upper bound of complexity $O( (p + q)^2)$.  We also show that the greedy algorithm for gCCA guarantees the full recovery of true related and irrelevant variables for associations between two high-dimensional datasets with a high probability under mild assumptions. In addition, we demonstrate that the greedy algorithm for gCCA has square-root estimation consistency for the canonical correlation estimation. To the best of our knowledge, non-asymptotic analysis of square-root estimation consistency in CCA has not been explored, although some asymptotic analyses have been conducted in this area \citep{anderson1999asymptotic}.

%This is aligned with our theoretical results.   %Our analysis techniques can be applied to analogous problems related to the best subset selection due to reasonable and inclusive assumptions.
The simulation studies validate our theoretical results by showing that gCCA outperforms conventional methods to more accurately reveal the correlated variables and reduce the estimation bias of the canonical correlation. In our data application, we use gCCA to identify the systematic correlations between two DNA methylation pathways and two RNA expression pathways with both negative and positive correlations, revealing the new interactive biological pathways and their associations using multi-omics data.

\section*{Acknowledgement}

This research was funded by the National Institute on Drug Abuse of the National Institutes of Health under Award Number 1DP1DA04896801. %Additional support for computer cluster was provided by NIH R01 grants EB008432 and EB008281.

\section*{Availability and Implementation}

The R code that implements gCCA is available at \href{https://github.com/hjpark0820/gCCA}{https://github.com/hjpark0820/gCCA}.

%An interesting topic of prospective research includes proposing theoretical results for the correlation matrix with multiple subgraphs. In addition, we can extend the present proof to the setup under the existence of interconnected structures, where subgraphs have associations with each other. Lastly, studying algorithms based on better detection schemes than Greedy algorithm can be another variant of interest for future work. 

%  The \backmatter command formats the subsequent headings so that they
%  are in the journal style.  Please keep this command in your document
%  in this position, right after the final section of the main part of 
%  the paper and right before the Acknowledgements, Supplementary Materials,
%  and References sections. 

%  This section is optional.  Here is where you will want to cite
%  grants, people who helped with the paper, etc.  But keep it short!

%  If your paper refers to supplementary web material, then you MUST
%  include this section!!  See Instructions for Authors at the journal
%  website http://www.biometrics.tibs.org

\bibliographystyle{biom}
\bibliography{biomsample}

\section{Supplementary Materials}
\label{sec:supp}
\subsection{Proof of the statement in Assumption \ref{ass:sub} for Gaussian distributions}
\label{ssec:ass1}
Let $f_n(r|\rho)$ be the density for a sample correlation $R$ of bivariate normal distribution with correlation coefficient $\rho$ and sample size $n>3$. In the work of \cite{hotelling1953new}, it is written as follows:
\begin{eqnarray*}
    f_n(r|\rho) = \frac{n-1}{\sqrt{2\pi}} \frac{\Gamma(n)}{\Gamma(n+\frac{1}{2})} (1-\rho^2)^{\frac{n}{2}}(1-r^2)^{\frac{n-3}{2}}(1-\rho r)^{-n+\frac{1}{2}} \prescript{~}{2}F_1\left(\frac{1}{2},\frac{1}{2},n+\frac{1}{2},\frac{1+r\rho}{2}\right),
\end{eqnarray*}
where $\prescript{}{2}F_1$ is the hypergeometric function such that
\begin{eqnarray*}
\prescript{}{2}F_1(a,b,c,x) =  1+ \frac{ab}{c}x + \frac{a(a+1)b(b+1)}{2!c(c+1)}x^2 + \dots,  
\end{eqnarray*}
which is convergent for $x\in(-1,1)$. Without loss of generality, we can consider $0\leq \rho<1$ as the argument for the case $\rho < 0$ is symmetric for the case $\rho >0$. For $|\rho|=1$, the upper bound is trivial. To find an upper bound of $P(|R-\rho|>\epsilon)$ for $\epsilon\geq 0$, we consider three mutually exclusive cases: (i) $R-\rho>\epsilon$, (ii) $-(R-\rho)<\epsilon$ and $0<\epsilon<\rho$, and (iii) $-(R-\rho)<\epsilon$ and $\rho<\epsilon$. First, we evaluate $P(R-\rho>\epsilon)$ for case (i). 
\begin{eqnarray*}
    &&P(R-\rho>\epsilon) \\
    &=& \int_{\rho+\epsilon}^1 \frac{n-1}{\sqrt{2\pi}} \frac{\Gamma(n)}{\Gamma(n+\frac{1}{2})} (1-\rho^2)^{\frac{n}{2}}(1-r^2)^{\frac{n-3}{2}}(1-\rho r)^{-n+\frac{1}{2}}\prescript{}{2}F_1\left(\frac{1}{2},\frac{1}{2},n+\frac{1}{2},\frac{1+r\rho}{2}\right) dr\\
    &\leq & M_1\frac{n-1}{\sqrt{2\pi}} \frac{\Gamma(n)}{\Gamma(n+\frac{1}{2})}  \int_{\rho+\epsilon}^1 (1-\rho^2)^{\frac{n}{2}}(1-r^2)^{\frac{n-3}{2}}(1-\rho r)^{-n+\frac{1}{2}} dr
\end{eqnarray*}
where $M_1 = \sup_{r\in[-1,1]}\prescript{}{2}F_1(1/2,1/2,n+1/2,\frac{1+r\rho}{2})< \infty$. Note that $\prescript{}{2}F_1\left(\frac{1}{2},\frac{1}{2},n+\frac{1}{2},\frac{1+r\rho}{2}\right) $ is positive as $0<\frac{1+r\rho}{2} < 1$ and $\prescript{}{2}F_1\left(\frac{1}{2},\frac{1}{2},n+\frac{1}{2},\frac{1+r\rho}{2}\right) \leq \prescript{}{2}F_1\left(\frac{1}{2},\frac{1}{2},\frac{3}{2},\frac{1+r\rho}{2}\right) = \text{arcsin}\left(\sqrt{\frac{1+r\rho}{2}}\right)/\sqrt{\frac{1+r\rho}{2}}$ \citep{lozier2003nist}. As $\text{arcsin}\left(\sqrt{\frac{1+r\rho}{2}}\right)/\sqrt{\frac{1+r\rho}{2}}$ is continuous in $r\in[-1,1]$, $\prescript{}{2}F_1\left(\frac{1}{2},\frac{1}{2},\frac{3}{2},\frac{1+r\rho}{2}\right)$ is bounded by a positive constant $M_1$.

Using $(a+b)/2 \geq \sqrt{ab}$ for $a,b>0$, we have
\begin{eqnarray*}
(1-\rho^2)^{\frac{n-3}{2}}(1-r^2)^{\frac{n-3}{2}}(1-\rho r)^{-(n-3)}
    &=& \left(\left(\frac{1-\rho^2}{1-\rho r}\right)^{1/2}\left(\frac{1-r^2}{1-\rho r}\right)^{1/2}\right)^{n-3}\\
    &\leq& \left(\frac{1}{2}\cdot\frac{(1-\rho^2) + 1-r^2}{1-\rho r}\right)^{n-3} \\
    &=&  \left(1 - \frac{1}{2}\cdot\frac{(r-\rho)^2}{1-\rho r}\right)^{n-3}.
\end{eqnarray*}

Accordingly, we have
\begin{eqnarray*}
    && \int_{\rho+\epsilon}^1 (1-\rho^2)^{\frac{n}{2}}(1-r^2)^{\frac{n-3}{2}}(1-\rho r)^{-n+\frac{1}{2}} dr\\
    &\leq& (1-\rho^2)^{\frac{3}{2}}(1-\rho)^{-\frac{5}{2}} \int_{\rho+\epsilon}^1 (1-\rho^2)^{\frac{n-3}{2}}(1-r^2)^{\frac{n-3}{2}} (1-\rho r)^{-n+3}  dr\\
    &\leq& (1-\rho^2)^{\frac{3}{2}}(1-\rho)^{-\frac{5}{2}} \int_{\rho+\epsilon}^1 \left(1 - \frac{1}{2}\cdot\frac{(r-\rho)^2}{1-\rho r}\right)^{n-3} dr\\
\end{eqnarray*}

Using $1-x \leq e^{-x}$ for $x>0$, for $r>\rho+\epsilon$, we have
\begin{eqnarray*}
    \left(1 - \frac{1}{2}\cdot\frac{(r-\rho)^2}{1-\rho r}\right)^{n-3} \leq \exp\left(-\frac{(n-3)(r-\rho)^2}{2(1-\rho r)}\right)\leq \exp\left(-\frac{(n-3)(r-\rho)^2}{2(1-\rho^2 )}\right).
\end{eqnarray*}

Thus, we have
\begin{eqnarray*}
    \int_{\rho+\epsilon}^1 \left(1 - \frac{1}{2}\cdot\frac{(r-\rho)^2}{1-\rho r}\right)^{n-3} dr \leq \int_{\rho+\epsilon}^1 \exp\left(-\frac{(n-3)(r-\rho)^2}{2(1-\rho^2)}\right) dr \leq \int_{\epsilon}^\infty \exp\left(-\frac{(n-3)r^2}{2(1-\rho^2)}\right) dr.
\end{eqnarray*}

Using the inequality  $1-\Phi(z)\leq \exp(-z^2/2),~z>0$, where $\Phi(\cdot)$ is the CDF of standard normal distribution, we have 
\begin{eqnarray*}
    1-\Phi\left(\frac{\sqrt{n-3}\epsilon}{\sqrt{1-\rho^2}}\right) = \int_{\epsilon}^\infty \frac{\sqrt{n-3}}{\sqrt{2\pi (1-\rho^2)}} \exp\left(-\frac{(n-3)r^2}{2(1-\rho^2)}\right) dr \leq \exp\left(-\frac{(n-3)\epsilon^2}{2(1-\rho^2)} \right).
\end{eqnarray*}

\begin{eqnarray}
    P(R-\rho>\epsilon) &\leq&  M_1 \frac{n-1}{\sqrt{2\pi}} \frac{\Gamma(n)}{\Gamma(n+\frac{1}{2})} \frac{\sqrt{2\pi (1-\rho^2)}}{\sqrt{n-3}} (1-\rho^2)^{\frac{3}{2}}(1-\rho)^{-\frac{5}{2}}\exp\left(-\frac{n\epsilon^2}{2(1-\rho^2)} \right)\nonumber\\
    &\leq& \frac{(n-1)\Gamma(n)}{\sqrt{n-3}\Gamma(n+\frac{1}{2})}  (1-\rho^2)^{2}(1-\rho)^{-\frac{5}{2}}\exp\left(-\frac{(n-3)\epsilon^2}{2(1-\rho^2)} \right)\nonumber\\
    &\leq&  M_1  M_2 (1+\rho)^2(1-\rho)^{-\frac{1}{2}}\exp\left(-\frac{(n-3)\epsilon^2}{2(1-\rho^2)} \right),\label{eq:case1}
\end{eqnarray}
where $M_2 = \sup_{n\geq 4} \frac{(n-1)\Gamma(n)}{\sqrt{n-3}\Gamma(n+\frac{1}{2})}$. $M_2$ is finite as $\frac{\Gamma(n)}{\Gamma(n+\frac{1}{2})} \leq \sqrt{\frac{1}{n+1/4}}$ \citep{qi2010bounds}.

Next, we calculate $P( -(R - \rho) >  \epsilon)$ for $0<\epsilon<\rho$ for case (ii). Note that
\begin{eqnarray*}
    && \int_{0}^{\rho-\epsilon} f_n(r|\rho) dr \\
    &\leq& \int_{0}^{\rho-\epsilon} \frac{n-1}{\sqrt{2\pi}} \frac{\Gamma(n)}{\Gamma(n+\frac{1}{2})} (1-\rho^2)^{\frac{n}{2}}(1-r^2)^{\frac{n-3}{2}}(1-\rho r)^{-n+\frac{1}{2}}\prescript{}{2}F_1\left(\frac{1}{2},\frac{1}{2},n+\frac{1}{2},\frac{1+r\rho}{2}\right) dr.
\end{eqnarray*}

Similarly to the proof for case (i), we have
\begin{eqnarray*}
    && \int_{0}^{\rho-\epsilon} (1-\rho^2)^{\frac{n}{2}}(1-r^2)^{\frac{n-3}{2}}(1-\rho r)^{-n+\frac{1}{2}} dr\\
    &\leq& (1-\rho^2)^{\frac{3}{2}}(1-\rho^2)^{-\frac{5}{2}} \int_{0}^{\rho-\epsilon} (1-\rho^2)^{\frac{n-3}{2}}(1-r^2)^{\frac{n-3}{2}} (1-\rho r)^{-n+3}  dr\\
    &\leq& (1-\rho^2)^{-1} \int_{0}^{\rho-\epsilon} \left(1 - \frac{1}{2}\cdot\frac{(r-\rho)^2}{1-\rho r}\right)^{n-3} dr\\
    &\leq& (1-\rho^2)^{-1} \int_{0}^{\rho-\epsilon} \left(1 - \frac{1}{2}(r-\rho)^2\right)^{n-3} dr\\
    &\leq& (1-\rho^2)^{-1} \int_{0}^{\rho-\epsilon} \exp\left(-\frac{(n-3)(r-\rho)^2}{2}\right) dr
\end{eqnarray*}

Accordingly, we have

\begin{eqnarray*}
    1-\Phi\left(\sqrt{n-3}\epsilon\right) = \int_{-\infty}^{\rho-\epsilon} \frac{\sqrt{n-3}}{\sqrt{2\pi}} \exp\left(-\frac{(n-3)(r-\rho)^2}{2}\right) dr \leq \exp\left(-\frac{(n-3)\epsilon^2}{2} \right).
\end{eqnarray*}

Thus, we have

\begin{eqnarray}
    &&\int_{0}^{\rho-\epsilon} \frac{n-1}{\sqrt{2\pi}} \frac{\Gamma(n)}{\Gamma(n+\frac{1}{2})} (1-\rho^2)^{\frac{n}{2}}(1-r^2)^{\frac{n-3}{2}}(1-\rho r)^{-n+\frac{1}{2}} \prescript{}{2}F_1\left(\frac{1}{2},\frac{1}{2},n+\frac{1}{2},\frac{1+r\rho}{2}\right)  dr\nonumber \\
    &\leq&  M_1 \frac{n-1}{\sqrt{2\pi}} \frac{\Gamma(n)}{\Gamma(n+\frac{1}{2})} \frac{\sqrt{2\pi}}{\sqrt{n-3}} (1-\rho^2)^{-1}\exp\left(-\frac{(n-3)\epsilon^2}{2} \right)\nonumber\\
    &\leq&  M_1 M_2 (1-\rho^2)^{-1}\exp\left(-\frac{(n-3)\epsilon^2}{2} \right).\label{eq:case2}
\end{eqnarray}

Lastly, we calculate $P( R - \rho <  - \epsilon)$ for $\rho<\epsilon$. Based on a similar logic to those in the previous two cases, we have
\begin{eqnarray*}
    && \int_{-1}^{\rho-\epsilon} (1-\rho^2)^{\frac{n}{2}}(1-r^2)^{\frac{n-3}{2}}(1-\rho r)^{-n+\frac{1}{2}} dr\\
    &\leq& (1-\rho^2)^{\frac{3}{2}} \int_{-1}^{\rho-\epsilon} (1-\rho^2)^{\frac{n-3}{2}}(1-r^2)^{\frac{n-3}{2}} (1-\rho r)^{-n+3}  dr\\
    &\leq& (1-\rho^2)^{\frac{3}{2}} \int_{-1}^{\rho-\epsilon} \left(1 - \frac{1}{2}\cdot\frac{(r-\rho)^2}{1-\rho r}\right)^{n-3} dr\\
    &\leq& (1-\rho^2)^{\frac{3}{2}}  \int_{-1}^{\rho-\epsilon} \left(1 - \frac{(r-\rho)^2}{2(1+\rho)}\right)^{n-3} dr\\
    &\leq& (1-\rho^2)^{\frac{3}{2}}\int_{-1}^{\rho-\epsilon} \exp\left(-\frac{(n-3)(r-\rho)^2}{2(1+\rho)}\right) dr
\end{eqnarray*}

Accordingly, we have

\begin{eqnarray*}
    1-\Phi\left(\frac{\sqrt{n-3}\epsilon}{\sqrt{1+\rho}}\right) = \int_{-\infty}^{\rho-\epsilon} \frac{\sqrt{n-3}}{\sqrt{2\pi(1+\rho)}} \exp\left(-\frac{(n-3)(r-\rho)^2}{2(1+\rho)}\right) dr \leq \exp\left(-\frac{(n-3)\epsilon^2}{2(1+\rho)} \right).
\end{eqnarray*}

Thus, we have

\begin{eqnarray}
    &&\int_{-1}^{\rho-\epsilon} \frac{n-1}{\sqrt{2\pi}} \frac{\Gamma(n)}{\Gamma(n+\frac{1}{2})} (1-\rho^2)^{\frac{n}{2}}(1-r^2)^{\frac{n-3}{2}}(1-\rho r)^{-n+\frac{1}{2}} \prescript{}{2}F_1\left(\frac{1}{2},\frac{1}{2},n+\frac{1}{2},\frac{1+r\rho}{2}\right)   dr \nonumber\\
    &\leq&  M_1 \frac{n-1}{\sqrt{2\pi}} \frac{\Gamma(n)}{\Gamma(n+\frac{1}{2})} \frac{\sqrt{2\pi(1+\rho)}}{\sqrt{n-3}} (1-\rho^2)^{\frac{3}{2}} \exp\left(-\frac{(n-3)\epsilon^2}{2(1+\rho)} \right)\nonumber\\
    &\leq&  M_1 M_2  (1+\rho)^{\frac{1}{2}}\exp\left(-\frac{(n-3)\epsilon^2}{2(1+\rho)} \right).\label{eq:case3}
\end{eqnarray}

Putting \eqref{eq:case1}, \eqref{eq:case2}, and \eqref{eq:case3} all together, we have
\begin{eqnarray*}
    P(|R-\rho|>\epsilon ) \leq 2 M_1 M_2(1-\rho)^{-1}  \exp\left(-\frac{(n-3)\epsilon^2}{2(1+\rho)} \right).
\end{eqnarray*}
Therefore, there exist $s_1>0$ and $s_2>0$ such that 
\begin{eqnarray*}
     P(|R-\rho|>\epsilon )  \leq s_1 \exp\left(-\frac{n\epsilon^2}{2s_2} \right).
\end{eqnarray*}

\subsection{Proof of the equivalence of Assumption \ref{ass:sub} to $E[e^{\lambda (R_{ij}-\theta_{ij})}] \leq s_3 e^{s_4^2\lambda^2/n}$}
\label{ssec:ass11}
Let $X=\sqrt{n}(R_{ij}-\theta_{ij})/(\sqrt{2}s_2)$ for ease of presentation. Then, we have
\begin{eqnarray*}
    E\left[\left|X\right|^p\right] &=& \int_{0}^\infty P(|X|^p \geq u) du\\
    &=& \int_{0}^\infty P(|X| \geq t)pt^{p-1} dt\\
    &\leq& \int_{0}^\infty s_1 e^{-t^2} pt^{p-1} dt.
\end{eqnarray*}
By letting $t^2=s$ and then using the definition of the Gamma function, we have 
\begin{eqnarray*}
\int_{0}^\infty s_1 e^{-t^2} pt^{p-1} dt  &=& (s_1/2)p\Gamma(p/2).
\end{eqnarray*}
Then, with the Stirling approximation $\Gamma(x) \leq x^x$, we get
\begin{eqnarray*}
    E\left[\left|X\right|^p\right]\leq (s_1/2)(p/2)^{p/2}.
\end{eqnarray*}
Using $Ee^{\lambda^2 X^2} = E(1+\sum_{p=1}^\infty \frac{(\lambda^2 X^2)^p}{p!} ) \leq 1+\sum_{p=1}^\infty \frac{\lambda^{2p}E[X^{2p}]}{p!}$ and $p! \geq
(p/e)^p$, we have
\begin{eqnarray*}
    Ee^{\lambda^2X^2} \leq  1+ \sum_{p=1}^\infty \frac{(s_1/2)(2\lambda^2 p)^p}{(p/2)^p} \leq \frac{s_1}{2} \sum_{p=0}^\infty (2e\lambda^2)^p = \frac{s_1}{2} \cdot \frac{1}{1-2e \lambda^2},
\end{eqnarray*}
provided that $2e\lambda^2 < 1$. Using $1/(1-x)\leq e^{2x}$ for $x\in [0,1/2]$, we have
\begin{eqnarray*}
    Ee^{\lambda^2X^2} \leq \frac{s_1}{2} e^{4e\lambda^2},~|\lambda|\leq \frac{1}{\sqrt{2e}}.
\end{eqnarray*}

Now, we focus on $Ee^{\lambda X}$. For $|\lambda| \leq 1$, using $e^x \leq x+ e^{x^2}$ and $E[X]=0$, we have
\begin{eqnarray*}
Ee^{\lambda X} \leq E(\lambda X + e^{\lambda^2 X^2}) = Ee^{\lambda^2 X^2} \leq \frac{s_1}{2} e^{4e\lambda^2}.
\end{eqnarray*}

Next, we prove the case with $|\lambda|\geq 1$. Using $2\lambda x \leq \lambda^2 +x^2$, we have
\begin{eqnarray*}
    E e^{\lambda X} \leq e^{\lambda^2/2} Ee^{X^2/2} \leq (s_1/2) e^{\lambda^2/2} e^{2e} \leq (s_1/2) e^{2e} e^{\lambda^2},~~~\text{for}~|\lambda|\geq 1.
\end{eqnarray*}

Putting the two cases $|\lambda|\leq 1$ and $|\lambda|\geq 1$ together, for $\lambda\in \mathbb{R}^1$,  we have
\begin{eqnarray*}
     E e^{\lambda X} \leq (s_1/2)e^{2e} e^{4e\lambda^2}.
\end{eqnarray*}

Letting $X=\sqrt{n}(R_{ij}-\theta_{ij})/(\sqrt{2}s_2)$, we have
\begin{eqnarray*}
     E e^{\lambda (R_{ij}-\theta_{ij})} \leq (s_1/2)e^{2e} e^{2es_2^2 \lambda^2/n}.
\end{eqnarray*}

Thus, by defining $s_3=(s_1/2)e^{2e}$ and $s_4^2 = 2e s_2^2$, we have
\begin{eqnarray}
     E e^{\lambda (R_{ij}-\theta_{ij})} \leq s_3 e^{s_4^2 \lambda^2/n}. \label{eq:ass2}
\end{eqnarray}

Now, we show that there exist $s_1,s_2>0$ such that $P(|R_{ij}-\theta_{ij}|>t) \leq s_1\exp(-\frac{t^2}{2s_2^2})$, if \eqref{eq:ass2} is true. Note that
\begin{eqnarray*}
    P(R_{ij}-\theta_{ij}\geq t) = P(e^{\lambda (R_{ij}-\theta_{ij})} \geq e^{\lambda t}) \leq e^{-\lambda t}Ee^{\lambda (R_{ij}-\theta_{ij})} \leq e^{-\lambda t}s_3e^{s_4^2 \lambda^2 }= s_3e^{-\lambda t + s_4^2 \lambda^2}.
\end{eqnarray*}
Thus, by letting $\lambda=t/2s_4^2$, we have
\begin{eqnarray*}
    P(R_{ij}-\theta_{ij} \geq t)\leq s_3 e^{-t^2/4s_4^2}.
\end{eqnarray*}
We can get the same result for $P(R_{ij}-\theta_{ij} < -t)$ for $t<0$ with the same logic. Therefore, we have
\begin{eqnarray*}
    P(|R_{ij}-\theta_{ij}| \geq t)\leq 2s_3 e^{-nt^2/4s_4^2}.
\end{eqnarray*}

\subsection{Proof of the statement in Assumption \ref{ass:ass2} for Gaussian distributions}
\label{ssec:ass2}
Let ${\bf X}_i = (X_{i1},X_{i2},\dots,X_{in})^\top$ be a random vector such that each $X_{ij}$ is independently generated from $N(\mu_i,\sigma_i^2)$. We assume that ${\bf X}_1$, ${\bf X}_2$, and ${\bf X}_3$ are mutually independent. We let centered and standardized vectors of ${\bf X}_i$, ${\bf X}_2$, and ${\bf X}_3$, with sample size $n$, denoted by ${\bf Y}_1$, ${\bf Y}_2$, and ${\bf Y}_3$, respectively. Since they are centered and standardized, they are in $\mathbb{S}^{d-1}$. In addition, $R_{ij} = {\bf Y}_i^\top {\bf Y}_j$. Because each random variable ${\bf X}_i$ is independent from each other, ${\bf Y}_i$ is unformly distributed in $\mathbb{S}^{d-1}$. Consider the joint density of ${\bf Y}_1$ and $R_{12}=r_{12}$. Then, we have
\begin{eqnarray*}
    f({\bf Y}_1=y_1,R_{12}=r_{12})=f(R_{12}=r_{12}|{\bf Y}_1=y_1)f({\bf Y}_1=y_1).
\end{eqnarray*}
Here, note that $R_{12}=r_{12}$ represents ${\bf Y}_2^\top y_1 = r_{12}$ and that $r_{12}$ represents $\text{cos}(\theta_{12})$, where $\theta_{12}$ is the angle between ${\bf Y}_1$ and ${\bf Y}_2$. The set $\{y:y^\top y_1 = r_{12}\} \in \mathbb{S}^{d-1}$ is the collection of vectors in $\mathbb{S}^{d-1}$ that have the angle $\theta_{12} = \text{cos}^{-1}(r_{12})$ with $y_1$. Since ${\bf Y}_2$ is independent of ${\bf Y}_1$ and uniformly distributed over $\mathbb{S}^{d-1}$, the probability density of $\{y:y^\top y_1 = r_{12}\}$ does not depend on the value of $y_1$. This means 
\begin{eqnarray*}
    f(R_{12}=r_{12}|{\bf Y}_1=y_1) = f(R_{12}=r_{12}).
\end{eqnarray*}
Accordingly, we have
\begin{eqnarray*}
    f({\bf Y}_1=y_1,R_{12}=r_{12})=f(R_{12}=r_{12})f({\bf Y}_1=y_1).
\end{eqnarray*}
Thus, ${\bf Y}_1$ and $R_{12}$ are indenpendent. Because ${\bf Y}_3$ is also independent from $R_{12}$, $R_{13}={\bf Y}_1^\top {\bf Y}_3$ and $R_{12}$ are independent. Therefore, Assumption \ref{ass:ass2} holds for normal distributions.

\subsection{Proof of Lemma \ref{lem:1}}

\begin{proof}
    Let $\frac{\delta}{pq} = s_1\exp(-na^2/2s_2^2)$ in Assumption \ref{ass:sub}. Then, we have $P\left(|R_{ij}-\rho_{ij}| > \sqrt{\frac{2s_2^2}{n}\log \left(\frac{s_1pq}{\delta}\right)}\right) \leq \frac{\delta}{pq}$. Thus, we have
    \begin{eqnarray*}
        P\left(|R_{ij}-\rho_{ij}| > \sqrt{\frac{2s_2^2}{n}\log \left(\frac{s_1pq}{\delta}\right)},~\text{for~all}~(i,j)\in [p]\otimes [q]\right) \leq pq\frac{\delta}{pq} = \delta.
    \end{eqnarray*} %\hfill$\square$
\end{proof}

\subsection{Proof of Lemma \ref{lem:2}}

\begin{proof}
By Lemma \ref{lem:1}, for all $i$ and $j$, we have $|R_{ij}-\rho_{ij}| < \sqrt{\frac{2s_2^2}{n}\log \left(\frac{s_1pq}{\delta}\right)}$ with probability at least $1-\delta$. If $n\geq (\eta/\rho)^2 (2s_2^2\log \left(s_1pq/\delta\right))$, we have $\sqrt{\frac{2s_2^2}{n}\log \left(\frac{s_1pq}{\delta}\right)} < \rho/\eta.$ Accordingly, we have $|R_{ij}|> \frac{\eta-1}{\eta}\rho$ for $i\in I_{X}~\text{and}~j\in I_{Y}$ and $|R_{ij}| < \frac{1}{\eta}\rho$ for $i\in I_{X}^c~\text{or}~j\in I_{Y}^c$. Now, we consider $|R_{ij}^\varepsilon|$ for $0\leq \varepsilon < \rho/2$. As $\frac{\eta-1}{\eta}\rho>\varepsilon$ and $|R_{ij}|\geq |R_{ij}^\varepsilon|$, we still have 
\begin{eqnarray}
    |R_{ij}^\varepsilon|> \frac{\eta-1}{\eta}\rho,~~~\text{for}~i\in I_{X}~\text{and}~j\in I_{Y}\label{eq:r23}
\end{eqnarray}
and
\begin{eqnarray}
    |R_{i'j}^\varepsilon| < \frac{1}{\eta}\rho,~~~\text{for}~i'\in I_{X}^c~\text{or}~j\in I_{Y}^c.\label{eq:r13}
\end{eqnarray}

Thus, we have $\sum_{j\in I_Y} |R_{ij}^\varepsilon| > q_0((\eta-1)/\eta)\rho$ for all $i\in I_X$ and $\sum_{j\in I_Y} |R_{i'j}^\varepsilon| < q_0(\rho/\eta)$ for all $i'\in I_X^c$ with probability at least $1-\delta$. Accordingly, we have $\sum_{j\in I_Y} (|R_{ij}^\varepsilon|  - |R_{i'j}^\varepsilon|) > q_0\rho(\eta-2)/\eta$ for $i\in I_X$ and $i'\in I_X^c$  with probability at least $1-\delta$. In the same way, we can show
$\sum_{i\in I_X} (|R_{ij}^\varepsilon|  - |R_{ij'}^\varepsilon|) > p_0\rho(\eta-2)/\eta$ for $j\in I_Y$ and $j'\in I_Y^c$ with probability at least $1-\delta$.
\end{proof}

\subsection{Proof of Lemma \ref{lem:3}}

\begin{proof}

 Assumption \ref{ass:ass2} is applicable to $(i_1,h_1(t))$ and $(i_2,h_1(t))$ for all  $t \in [q-q_0]$, as $(i_1,h_1(t))$ and $(i_2,h_1(t))$ are not in the subgraph for all $t \in [q-q_0]$. Consequently, based on Assumption \ref{ass:sub}, for all $a>0$, we have
\begin{eqnarray*}
    E\left[e^{a(|R_{i_1h_1(t)}^\varepsilon| - |R_{i_1h_1(t)}^\varepsilon|) }\right]=E\left[e^{a|R_{i_1h_1(t)}^\varepsilon|}\right] E\left[e^{-a|R_{i_1h_1(t)}^\varepsilon| }\right]  \leq s_3 e^{\frac{s_4^2a^2}{n}}.
\end{eqnarray*}
Letting $a=\sqrt{n/s_4^2}$, we have
\begin{eqnarray*}
    E\left[e^{\sqrt{\frac{n}{s_4^2}}(|R_{i_1h_1(t)}^\varepsilon| - |R_{i_1h_1(t)}^\varepsilon|) }\right]  \leq s_3 e.
\end{eqnarray*}

For $i_1,i_2\in [p]$, let 
\begin{eqnarray*}
M_t^{i_1i_2}= \exp\left(\sqrt{\frac{n}{s_4^2}}\sum_{j=1}^t(|R_{i_1h_1(j)}^\varepsilon| - |R_{i_2h_1(j)}^\varepsilon|)-(1+\log s_3)t\right)    
\end{eqnarray*}
    where $M_t^{i_1i_2} = M_{q-q_0}^{i_1i_2}$ for $t\geq q-q_0$ and $\tau$ be a stopping time with respect to the filtration $\{\mathcal{F}_{t-1}^{i_1i_2}\}_t$, where $\mathcal{F}_t^{i_1i_2} =\sigma\{ \{ |R_{i_1h_1(j)}^\varepsilon|,|R_{i_2h_1(j)}^\varepsilon|\}_{j=1}^t\}$. First, we claim $\{M_t^{i_1i_2}\}_{t=1}^{q-q_0}$ is a supermartingale. Let $D_t^{i_1i_2} = \exp\left(\sqrt{\frac{n}{s_4^2}}(|R_{i_1h_1(t)}^\varepsilon| - |R_{i_2h_1(t)}^\varepsilon|)-(1+\log s_3)\right)$. 
By Assumption \ref{ass:ass2}, $D_t^{i_1i_2}$ and  $\mathcal{F}_{t-1}^{i_1i_2}$ are independent. Thus, we have
\begin{eqnarray*}
E[D_t^{i_1i_2}|\mathcal{F}_{t-1}^{i_1i_2}] \leq E\left[e^{\sqrt{\frac{n}{s_4^2}}(|R_{i_1h_1(t)}^\varepsilon|-(1+\log s_3)) }\right]\leq 1.
\end{eqnarray*}
Clearly, $D_t^{i_1i_2}$ is $\mathcal{F}_t^{i_1i_2}$-measurable, as is $M_t^{i_1i_2}$. Further, we have $E[M_t^{i_1i_2}|\mathcal{F}_{t-1}^{i_1i_2}]=D_1^{i_1i_2}\cdots D_{t-1}^{i_1i_2}E[D_t^{i_1i_2}|\mathcal{F}_{t-1}^{i_1i_2}] \leq M_{t-1}^{i_1i_2}$. This shows that $\{M_t^{i_1i_2}\}$ is a supermartingale. By the convergence theorem for nonnegative supermartingales \citep{doob1953stochastic}, $M_\infty^{i_1i_2}=\lim_{t\rightarrow \infty} M_t^{i_1i_2}$ is almost surely well-defined. Hence, $M_\tau^{i_1i_2}$ is well-defined. Next, let $Q_t^{i_1i_2} = M_{\min(\tau,t)}^{i_1i_2}$ be a stopped version of $\{M_t^{i_1i_2}\}$. By Fatou's lemma \citep{rudin1976principles}, we have
\begin{eqnarray*}
    E[\text{liminf}_{t\rightarrow \infty} Q_t^{i_1i_2}] \leq \text{liminf}_{t\rightarrow \infty}E[Q_t^{i_1i_2}] \leq 1.
\end{eqnarray*}
This shows that $E[M_\tau^{i_1i_2}]\leq 1$ holds. Lastly, from $E[M_\tau^{i_1i_2}]\leq 1$, we get
\begin{eqnarray*}
    &&P\left(\sqrt{\frac{n}{s_4^2}}\sum_{j=1}^\tau (|R_{i_1h_1(j)}^\varepsilon|- |R_{i_2h_1(j)}^\varepsilon|) -\tau(1+\log s_3)  > \log \delta^{-1}\right)
    =P\left(M_\tau^{i_1i_2}\delta^{-1} > 1 \right)\\ &\leq& E[M_\tau^{i_1i_2} \delta] \leq \delta.
\end{eqnarray*}
In other words, we have
\begin{eqnarray*}
    \sqrt{\frac{n}{s_4^2}}\sum_{j=1}^\tau (|R_{i_1h_1(j)}^\varepsilon| - |R_{i_2h_1(j)}^\varepsilon|) -\tau(1+\log s_3)  > \log \delta^{-1}
\end{eqnarray*}
with probability at least $1-\delta$. Similarly, for $j_1\in I_Y$ and $j_2\in I_Y^c$, we have
\begin{eqnarray*}
    \sqrt{\frac{n}{s_4^2}}\sum_{t=1}^\tau (|R_{h_2(t)j_1}^\varepsilon| - |R_{h_2(t)j_2}^\varepsilon| ) -\tau (1+\log s_3) > \log \delta^{-1}
\end{eqnarray*}
with probability at least $1-\delta$,  where $h_2$ is a bijective function from $\{1,2,\dots,p-p_0\}$ to $I_X^c$.  By plugging $\frac{\delta}{2\max(p,q)^2}$ into $\delta$, if $n > \frac{\eta^2 s_4^2}{2\min(p_0,q_0)^2(\eta-2)^2\rho^2}\left( \max(p,q)(1+\log s_3) + \log \frac{2\max(p,q)^2}{\delta}\right)^2$, we have
    $$\sum_{t=1}^{\tau} (|R_{i_1h_1(t)}^\varepsilon| - |R_{i_2h_1(t)}^\varepsilon| ) \leq \frac{\min(p_0,q_0)(\eta-2)\rho}{\eta},$$
for all $i_1\in I_X$ and all $i_2\in I_X^c$ and 
    $$\sum_{t=1}^{\tau} (|R_{h_2(t)j_1}^\varepsilon| - |R_{h_2(t)j_2}^\varepsilon| ) \leq \frac{\min(p_0,q_0)(\eta-2)\rho}{\eta},$$
for all $j_1\in I_Y$ and all $j_2\in I_Y^c$ with probability at least $1-\delta$.
\end{proof} 
%  Here, we create the bibliographic entries manually, following the
%  journal style.  If you use this method or use natbib, PLEASE PAY
%  CAREFUL ATTENTION TO THE BIBLIOGRAPHIC STYLE IN A RECENT ISSUE OF
%  THE JOURNAL AND FOLLOW IT!  Failure to follow stylistic conventions
%  just lengthens the time spend copyediting your paper and hence its
%  position in the publication queue should it be accepted.

%  We greatly prefer that you incorporate the references for your
%  article into the body of the article as we have done here 
%  (you can use natbib or not as you choose) than use BiBTeX,
%  so that your article is self-contained in one file.
%  If you do use BiBTeX, please use the .bst file that comes with 
%  the distribution.

%\appendix

%  To get the journal style of heading for an appendix, mimic the following.

\label{lastpage}

\end{document}